%% file: arxiv_version.tex
\documentclass[letterpaper]{article} 
\usepackage{aaai24}  
\usepackage{times}  
\usepackage{helvet}  
\usepackage{courier}  
\usepackage[hyphens]{url}  
\usepackage{graphicx} 
\usepackage{amsmath}
\usepackage{amssymb}
\usepackage{hyperref} 
\usepackage{natbib}  
\usepackage{caption} 
\usepackage{xspace}
\usepackage{caption}
\usepackage{subcaption}
\usepackage[colorinlistoftodos]{todonotes}
{}
{}
{}
{}
\usepackage{algorithm}
\usepackage{algorithmic}
\usepackage{soul}
\usepackage{newfloat}
\usepackage{listings}
\nocopyright

\DeclareCaptionStyle{ruled}{labelfont=normalfont,labelsep=colon,strut=off} 
\floatstyle{ruled}
\newfloat{listing}{tb}{lst}{}
\floatname{listing}{Listing}
\newcommand{\etal}{\textit{et al}.~}
\newcommand{\ie}{\textit{i}.\textit{e}.~}
\newcommand{\eg}{\textit{e}.\textit{g}.~}

\title{Advancing Image Retrieval with Few-Shot Learning and Relevance Feedback}
\author{
}

\author{
    Boaz Lerner, Nir Darshan, Rami Ben-Ari \thanks{Correspondence to: Rami Ben-Ari $<$ramib@originai.co$>$}
}
\affiliations{
    OriginAI, Israel
}

\usepackage{bibentry}

\begin{document}
\pagestyle{plain}

\maketitle

\begin{abstract}

With such a massive growth in the number of images stored, efficient search in a database has become a crucial endeavor managed by image retrieval systems. Image Retrieval with Relevance Feedback (IRRF) involves iterative human interaction during the retrieval process, yielding more meaningful outcomes. This process can be generally cast as a binary classification problem with only {\it few} labeled samples derived from user feedback. The IRRF task frames a unique few-shot learning characteristics including binary classification of imbalanced and asymmetric classes, all in an open-set regime. In this paper, we study this task through the lens of few-shot learning methods. 
We propose a new scheme based on a hyper-network, that is tailored to the task and facilitates swift adjustment to user feedback. Our approach's efficacy is validated through comprehensive evaluations on multiple benchmarks and two supplementary tasks, supported by theoretical analysis. We demonstrate the advantage of our model over strong baselines on 4 different datasets in IRRF, addressing also retrieval of images with multiple objects. Furthermore, we show that our method can attain SoTA results in few-shot one-class classification and reach comparable results in binary classification task of few-shot open-set recognition.
\end{abstract}
\section{Introduction}
\begin{figure*}[ht]
\centering
   \includegraphics[width=0.9\linewidth]{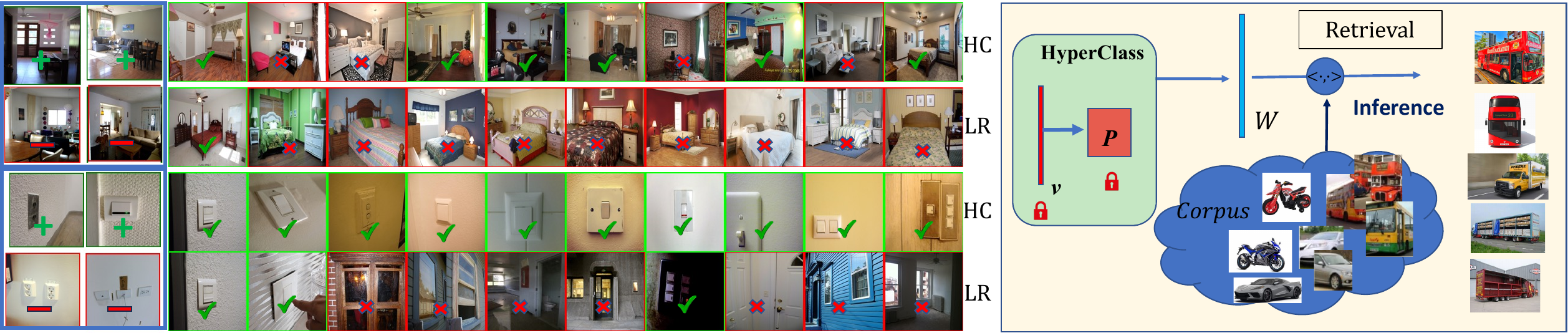} 
   \caption{{\bf Left:} The challenge in capturing the user intent. A case of two relevant and two irrelevant samples given by the user, on the left (separated by a blue frame). On the top, the query concept corresponds to a "Fan" on the ceiling, and at the bottom a "Light-switch". Top-k retrievals are shown (from left to right). HC (HyperClass) on the top is compared to LR (Logistic Regression) baseline on the bottom. Correct and incorrect retrievals are marked in green and red respectively. While LR is trapped by the gist of the image (the room in "Fan" and a frame in "Light-switch") HC better captures the correct object. {\bf Right:} Schematic view of our model usage for retrieval (after finetuning): Hyperclass generates $W$ for retrieval in a corpus.}
    \label{fig:Visual_and_InferenceScheme}
\end{figure*}

Image retrieval is a long-standing problem in computer vision commonly used in various applications such as e-commerce, social media, and digital asset management. Content-based Image Retrieval (CBIR) considers the image content for the retrieval task. 
It is often challenging to capture the user's intent from a single or few images, particularly when the images contain multiple objects in a complex scene. Image Retrieval with human Relevance Feedback (IRRF) \eg \cite{RelevanceFeedback_ICSPIS2017,putzu2020convolutional,medical_relevanceFeedback2018,Expertoslf2022} is an interactive method that refines the results iteratively, usually by having the user label a few retrieved images as {\it relevant} (positive-belongs to the query concept) or {\it irrelevant} (negative). These images are then added to the train set, or {\it support set} in FSL's terminology,  to improve the search results. The input of irrelevant samples (often hard negatives) allows discrimination between similar concepts or details in an image. Fig. \ref{fig:Visual_and_InferenceScheme} shows an example of the challenge in the retrieval task (our method HC vs. a baseline LR). On the top, the query corresponds to an image with a "Fan" on the ceiling (small object), and at the bottom a "Light-switch" is searched. Two relevant and two irrelevant (used to clarify the user's intent) user-tagged images are shown at the left (separated by blue frame). Top-k retrievals are presented (ranked from left to right). While a baseline (LR) is trapped by the gist of the image (the room instead of the "fan" and the frame instead of the "Light-switch"), our method HC better captures the semantic object derived from the user assignment of relevant and irrelevant samples.

The classification problem in IRRF is similar to few-shot
image classification but with several challenging and unique
characteristics. These include  imbalance, asymmetry and an open-set regime. Imbalance is exhibited as both the support set and the search corpus (test set) contain only few positive samples. Class asymmetry exists between the positive set, that comes from a homogeneous category (a single class) and the negative set, that is drawn from a complex multimodal distribution of many different classes. Finally, IRRF involves an open-set scenario where the number and type of negative classes/concepts in the train and test corpus are unknown.  

\begin{figure*}
    \centering
    \includegraphics[width=0.85\textwidth]{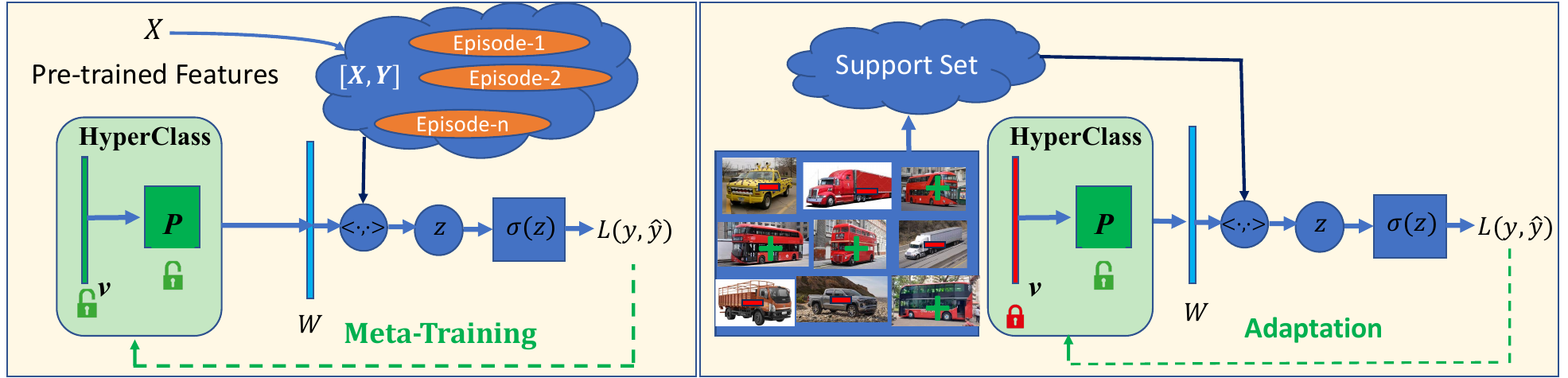}
    \caption{Our HyperClass trainig  scheme. {\bf Left: } Meta Training: Using pre-trained fixed features $X$, we learn a task-agnostic classifier $v$ and a projection head $P$, via MAML. The model outputs a linear classifier $W$, $z=W^T x$, $\sigma$ is the sigmoid function and $L$ is BCE loss. {\bf Right: } Test Adaptation: Given the user's tagged samples, HyperClass runs several adaptation steps to reach a new classifier for retrieval of a specific query concept.}
    \label{fig:HyperClass Scheme}
\end{figure*}

Despite the IRRF similarity to few-shot learning, previous methods have not explored its full potential, particularly {\it meta learning}. The study by \cite{putzu2020convolutional} can be related to a typical FSL strategy, of learning a backbone on closed set, image classification task, and then train a linear layer for {\it adaptation} in each IRRF cycle. Nonetheless, this approach is less suitable for the specific task attributes of IRRF. We tackle the previous shortcomings in two aspects. 
First, we suggest a Meta-Learning strategy, that mimics the exact problem characteristics. The second aspect refers to the design of our model and its training strategy (see Fig. \ref{fig:HyperClass Scheme} for a schematic view of the architecture). We decompose the standard linear classifier typically used in FSL to two components. A ''global" vector $v$ that represents a prior over a general task, and an adaptation matrix $P$,  adapting the few-shot classifier to a specific query concept, are both learned in MAML fashion. 

We further apply our approach to two more tasks that share similar characteristics, Few-shot One Class Classification (FSOCC) \cite{1way_proto,fs_occ-kozerawski,fs_occ-AAAI} (used for anomaly detection) and Few-shot Open-set Recognition (FSOR) ~\cite{FSOR_cvpr2020, FSOR_cvpr2021, FSOR_cvpr2022}. FSOCC attempts to learn a binary classifier with only few samples from one class. In Few-shot open-set recognition (FSOR), one has to deal with unseen classes, existing as distractors. The task requires an additional binary classification sub-task of detecting unknown out-of-set classes (non-support classes) appearing in the {\it query} set (for preliminary and background see suppl. material).

We evaluate our model on IRRF with realistic benchmarks, created from three different data sets and one popular few-shot benchmark, showing a significant improvement over prior art and strong baselines. We further present high performance on FSOCC and achieve competitive results on the binary FSOR subtask. Finally, we provide a theoretical analysis and justification to our method.

Overall, our contribution is three-fold:
1) We suggest a novel learning approach for IRRF, that builds upon ideas from few-shot learning. Our approach offers computationally efficient training, which can accommodate a variety of deep backbones.
2) Our approach can be further used for the demanding tasks of few-shot one-class classification (both inductive and transductive) and the subtask of in vs. out of set categorization in few-shot open-set learning.
3) We conduct a theoretical analysis revealing the advantage  of our method over common fine-tuning approaches in few shot learning.

\section{Related Work}     
Previous works address several aspects of the IRRF task. Some study the user interaction from the active-learning perspective ~\cite{ITAL2018, interactive_CBIR_2016, interactive_CBIR_2018}, aiming to maximize the retrieval results given selected samples. Another line of works address the problem of domain gap between the search and pre-trained domains \cite{Expertoslf2022} or adjustment of the IRRF to real world scenario to mimic human behavior \cite{lin2022block}. Yet, only few studies address the classification task, also considered as finding an equivalent query, \cite{IIR_CVIU02,RelevanceFeedback_ICSPIS2017,putzu2020convolutional,medical_relevanceFeedback2018,novelRF_neuroComputing2016}, which is the focus of this paper. \cite{OptimizedQueryIRRF2017,OptimizedQueryIRRF2018} suggest an IRRF method based on Rocchio's algorithm. Rocchio's algorithm enhances retrieval by adjusting the initial query vector based on modified distances from the relevant and irrelevant samples.
Others separate relevant and non-relevant images using Bayesian Networks \cite{BayesianNetworkClassifierIRRF}, CNN \cite{liu2017robust,pinjarkar2020deep,putzu2020convolutional}, Clustering \cite{dang2017multimodal}, Logistic Regression \cite{novelRF_neuroComputing2016} and Support Vector Machine \cite{SVM_IRRF2018,wang2016new}.  
In particular, Putzu \etal \cite{putzu2020convolutional} fine-tunes a linear layer over a pretrained CNN, a method equivalent to one of our baselines, which is inspired by FSL literature.  Nonetheless, this approach is less suitable for the specific task attributes of the IRRF task, which involves imbalanced, asymmetric, and open-set classification. In this context, we propose a few-shot learning training approach for IRRF. This model acquires prior knowledge on the IRRF task and adjusts to its distinctive attributes.
\section{Method}
\label{sec-method}
In this section we introduce the proposed approach, providing details on the model and the training scheme. Our model is a small hyper-network, dubbed HyperClass that outputs a linear binary classifier.  For a short background on FSL (and FSL terminology) we refer the reader to our suppl. material.

HyperClass seeks to benefit from the two main approaches of FSL: Transfer Learning and Meta Learning. We therefore divide the training into two stages. The first one aims to learn discriminative features, similar to \cite{GoodEmbedding}. Then, in the second stage, we apply a MAML approach on top of the pre-computed features. In contrast to common metric learning methods and MAML that adapt the embeddings in meta-train and/or in meta-test, we keep the embedding fixed, and train an adaptation module specialized to the specific embeddings. This adaptation module learns a task-agnostic classifier and a task-specific projection head adapting the classifier to a new task.
Figure \ref{fig:HyperClass Scheme} illustrates our training scheme and the associated components.

{\bf Feature Extraction:}
The goal of transfer learning is to learn a transferable embedding model $f_{\phi}$ which generalizes to a new task. To this end, we train a feature extractor on a standard multi-class task similar to ~\cite{GoodEmbedding}. This is depicted in Fig. \ref{fig:HyperClass Scheme}(a).

{\bf Adaptation Model:}
We start by describing our meta-training scheme. In our setting an image $I$ is mapped to a fixed representation $x=f_{\phi}(I)\in \mathbb{R}^d$ ($d$ - features dimension).
A straight forward and effective approach suggests learning a Logistic Regression (LR) classifier over the learned representations with the few samples from the support set, as in \cite{GoodEmbedding}. The LR parameters $\theta_l = \{W_l,b_l\}$ then include a weight term $W_l \in \mathbb{R}^{N \times d}$ (N-number of classes) and a bias term $b_l \in \mathbb{R}^d$ obtained by:
\begin{equation}
    \theta_l = {\arg\min_{W_l,b_l}} \sum_{i=1}^n \mathcal{L}^{bce}(W_lx_i^s+b_l,y_i^s) + \mathcal{R}(W_l,b_l),
    \label{eq:Linear Classifier}
\end{equation}
where $D^s=\{(x^{s}_{i}, y^{s}_{i})\}_{i=1}^n$ 
are the features and labels of the support set samples, $n$ denotes the number of support samples, and $\mathcal{L}^{bce}$ is the binary cross-entropy (BCE) loss. Finally, $\mathcal{R}(\cdot)$ is a regularization function.

We are also interested in a linear classifier due to scarcity of labeled samples and computational efficiency. However, we wish to leverage the information that variety of tasks has in common, and not to rely only on the task at hand. To this end, we learn a {\it prior}, represented by a {\it global} classifier and a good initialization for a projection head, adapting that classifier to a new task. The idea is to create an adapted classifier for each task with few samples and few gradient steps. This is performed with MAML optimization. We focus on binary classification, hence the classifier weights are reduced to $W \in \mathbb{R}^{d}$. More specifically, we introduce $v \in \mathbb{R}^d$ - a global {\it task-agnostic} classifier that is {\it projected} (transformed) into a subspace for each specific (local) task by: 
\begin{equation}
W=Pv+b     
\end{equation}
where $P \in \mathbb{R}^{d\times d}$ is the projection head and $b\in\mathbb{R}^d$ is a bias term. The prediction is then $\hat{y} = \sigma(W^Tx)$, with $\sigma$ standing for the Sigmoid function. 
For a schematic illustration of this part see Fig. \ref{fig:HyperClass Scheme}(b)). More specifically, our trainable parameters $\theta=\{v,P,b\}$ are updated with a few gradient steps on the support set of a given task $j$, with the following rule:

\begin{equation}
    \theta_j^{inner} = \theta - \nabla_{\theta}\mathcal{L}^{s}(\theta, D^s) + \mathcal{R}(\theta),
    \label{eq:Inner loop update}
\end{equation}

where $\mathcal{L}^{s}(\theta, D^s)$ refers to the BCE loss computed on predictions $\{\hat{y}^{s}_{i}\}_{i=1}^{n}$ and labels $\{y^{s}_{i}\}_{i=1}^{n}$. 
Then, in the outer loop we apply the adapted classifier on the query set for updating the global parameters, considering all inner loop updates, again with BCE loss $\mathcal{L}^{q}$, as follows:
\begin{equation}
    \theta^{outer} = \theta - \frac{1}{T}\sum_{j=1}^{T} \nabla_{\theta}\mathcal{L}^{q}(\theta_j^{inner}, D^q),
    \label{eq:Outer loop update}
\end{equation}
where $T$ is the number of meta-batch tasks and $D^q$ is the query set.

An outline of our meta-training pipeline is illustrated in Fig. \ref{fig:HyperClass Scheme}. The figure describes the meta-training and adaptation stages. In meta-training the model is trained over various task-episodes with both $v$ and $P$. In each cycle of IRRF, the vector $v$ remains fixed, while $P$ is updated (initialized from the meta-training process). The final retrieval scheme is shown in Fig. \ref{fig:Visual_and_InferenceScheme}. 

{\bf Training strategy for different use cases:}
In IRRF, we simulate the meta-testing scenario with binary, imbalanced and open-set tasks. Each task is formed by randomly selecting a positive class and extracting a small number of samples from that chosen class. Then, we consider all other classes as negatives and draw samples to create an imbalanced set. The query set follows the same strategy. In FSOCC and FSOR, only the support set is changed, to have few samples from {\it one} class (not binary). The query set remains the same, including both positive and negative samples. 
Note that in FSOCC task, there is a trivial solution (mode collapse) due to BCE loss trained on one class. We avoid this solution by taking the following actions, including negative samples in the query set, applying just a few adaptation steps and $L_2$ regularization. We also report results with transductive variant of our method which makes use of the query set features without the labels, during adaptation on the FSOCC task. In this setting the loss is a weighted combination of the BCE loss on the labeled set together with entropy minimization ~\cite{entropy_minimization} on the unlabeled query set. More specifically, In the transductive setting we make use of the few samples from the positive class together with the unlabeled query set which consists of both positive and negative samples. The difference to our standard training objective is that in each gradient update we take into account these two sets of samples. The labeled samples receive the same BCE loss, while the unlabeled samples are driven towards low entropy in the classifier prediction. This idea of Entropy Minimization is widely applied in the Semi-Supervised Learning literature, e.g. FixMatch. Formally, 
the transductive loss is computed as follows:
\begin{equation}
    \mathcal{L}^{transductive} = -\frac{1}{2n}\sum_{i=1}^{n} \log\hat{y}^{s}_{i} 
    -\frac{1}{2m}\sum_{i=1}^{m} \hat{y}^{q}_{i}\log\hat{y}^{q}_{i} 
    \label{eq:transductive loss}
\end{equation}
where $\{\hat{y}^{s}_{i}\}_{i=1}^{n}$ and $\{\hat{y}^{q}_{i}\}_{i=1}^{m}$ are the logits of the classifier for the support and query set respectively. We weigh the two losses equally.

Lastly, to handle FSOR, with multiple classes in meta-testing, we train a distinct projection head for every support class, utilizing solely samples derived from that specific class. For in-class probability, we take the max score over the trained classifiers. The meta-training model is the same as in FSOCC.  
\subsection{Theoretical Analysis}
\label{sec:Analysis}
Let us consider a binary classification task where the support samples are composed of the {\it positive} and {\it negative} sets: $x^s = x^s_p \cup x^s_n$, and we have a linear classifier $W \in \mathbb{R}^{d}$. Consider a simple application, where the classifier weights are set as the centroid of the positive support class. Hence, The classifier lies in a subspace spanned by the positive support feature vectors:
\begin{equation}
    W  = \frac{1}{|x^s_p|}\sum_{x\in x^s_p} x \in \mathbb{S}(x^s_p),
    \label{eq:Proto}
\end{equation}
where $\mathbb{S}(\cdot) $ denotes the {\it span} of a vector set. We refer to this classifier as {\it Proto}. Let us now consider an LR classifier previously used on top of a feature extractor in \cite{GoodEmbedding} (this is actually a simple linear layer with standard sigmoid function) and look at the resulting weight updates. The classifier change in gradient step $t$,  $\Delta W = W^{t+1} -  W^t$,   obtained by derivation of the loss in one gradient descent step (assuming unit learning rate w.l.o.g) is:
\begin{equation}
    \Delta W = \frac{1}{|x^s|}\sum_{x\in x^s_p} \alpha x - \frac{1}{|x^s|}\sum_{x\in x^s_n} (1-\alpha)x,
\end{equation}
where $\alpha:=1-\sigma(W^Tx) \in \mathbb{R}$ (see derivation in suppl. material). As expected, it holds that $W^t \in \mathbb{S}(x^s), \forall t \in T$. Therefore, once again the classifier lies in the sub-sapce spanned by the support set.

Next we derive the update scheme of the inner-loop for our HyperClass method. 
For sake of brevity, we show the term obtained for a positive $x^{+} \in x^s_p$ (for detailed derivation see the suppl. material):
\begin{align}
    \Delta W = 
    \alpha PP^Tx^{+}
    -\alpha (||v||_{2}^{2} + 1 - \alpha v^TP^Tx^{+})x^{+}
\end{align}
or briefly,
\begin{equation}
     \Delta W = Cx^{+} + \alpha PP^Tx^{+}
     ,
\end{equation}
with $C \in\mathbb{R}$ being a scalar value. This shows the impact of the projection head, learned in two phases, initialized via meta-training and fine-tuned during the adaption stage (see Fig. \ref{fig:HyperClass Scheme}). While the first component in $\Delta W$ lies in the support subspace as in LR, the additional term obtained from $P$, namely $\alpha PP^Tx^+$ extends the degrees of freedom, forming a richer hypothesis class (set of possible solutions) that can further be viewed as  ''tilting" the classifier in a direction out of the support {\it subspace} (see illustration in suppl. material). Our scheme can further be viewed as a generalization of \cite{GoodEmbedding}. HyperClass reduces to LR w.r.t. the direction of the update vector in a case where $P$ is an orthogonal matrix. Please see the visualization of the geometrical intuition in suppl. material. 

Deriving the adapted classifier after $k$ gradient steps, from the initial state ($W_0, P_0$) yields the following term (derivation provided in suppl. material): 
 \begin{align}
W_{k} = &W_0 + P_0P_0^T X^T \beta^1 + X^T \beta^2 + \beta^3 XX^TW_0 \nonumber \\  
& ~~~~~~+ XX^T(P_0P_0^T X^T) \beta^4,
\end{align}
denoted in matrix notation for sake of brevity. The vectors $\beta^1, \beta^2, \beta^4 \in \mathbb{R}^{|x^s|}$ are associated with the support samples (scalar for each sample) and $\beta^3 \in \mathbb{R}$ is a scalar. We observe that the two last terms are weighted by the sample covariance matrix $XX^T$ (assuming zero-centered features). This shows dependence of $W$ on the second order statistics, in contrast to LR, which utilizes only first order statistics. Deeper examination and derivation can be found in suppl. material.

An intuition behind the design of decomposing $W$ to $P$ and $v$ can also be attained from the Bayesian perspective. Our classifier in fact combines the learned prior knowledge (expressed by initial $v$ and $P$) with the evidence represented by the small labeled set, during adaptation.
\section{Experiments}
We evaluate our method comparing it to SoTA and strong baselines on three different tasks, IRRF, FSOCC and FSOR. In the following we elaborate on the different benchmarks. In all benchmarks, we start by training a feature extractor on a standard multi-class classification task similar to ~\cite{GoodEmbedding}.

\subsection{Image Retrieval with Relevance Feedback}
{\bf Datasets: }
We evaluate the performance of our method on IRRF task on four datasets;  Paris, Places, FSOD and MiniImagenet. Paris involves instance search, while Places is a semantic search, and both datasets present a cross-domain scenario, that exhibits a further challenge. FSOD introduces cross-domain object level retrieval and MiniImagenet reflects an in-domain retrieval case. Following ~\cite{PMF_CVPR22} we consider two protocols; one is more research-oriented and the other is more realistic and practical. 
The first protocol is evaluated on the FSL MiniImagenet ~\cite{miniImagenet} dataset. In order to create an IRRF  benchmark we split the data in the typical way to base/novel classes with train/val/test splits. The test set is used as the database for retrieval of the novel classes.

The more realistic protocol makes use of a strong backbone trained on a large external dataset, e.g. ImageNet ~\cite{imagenet}. For test we consider three different search concepts. 
For Paris (known as Paris-6K) benchmark we follow the standard protocol as suggested in~\cite{new_paris_oxford}. This dataset contains $11$ different monuments from Paris, plus 1M distractor images, from which we sampled a small subset, resulting in 9,994 images with 51-289 samples per-class and 8,204 distractors. {\bf Places} contains 365 different types of places. Our Places dataset consists of the validation set of Places365 \cite{places}, having 100 samples per-class with 30 randomly sampled classes used as queries. Lastly, we evaluate ourselves on object-level retrieval. In this case the query concept is often among several existing objects in the image, making the recognition a greater challenge. To this end, we built a new benchmark from the Few-Shot Object Detection (FSOD) dataset ~\cite{fsod}. FSOD dataset is split into base and novel classes. We used the novel set, containing ~14K images from 200 categories, for our benchmark. Again, 30 categories were sampled randomly as queries.  

Similar to \cite{RelevanceFeedback_ICSPIS2017,putzu2020convolutional,novelRF_neuroComputing2016} we evaluate the performance on the IRRF task by creating a pool of the top-k retrieval results from the trained classifier at each iteration, and call it the candidate set. We set a budget of 10 feedback samples received from the user. To emulate the human relevance feedback, we divide the candidate set into positive and negative set and randomly sample 80\% of the budget from the positive set and 20\% from the negative set. Testing our method with other composition of positive vs. negative sampling showed similar results (see results and discussion in suppl. material). Note that we opt for this selection strategy to evaluate the classification quality with small train sets. At each iteration, we add the selected samples with their corresponding label to the train set and train a new binary classifier for the next retrieval iteration. We then present the results as learning curves by mean average precision (mAP) and Precision@50 at each interaction step.

For all datasets we follow the same protocol: sample a user-query image from a certain class, consider all images belonging to that class (or containing the same object in FSOD) as relevant, while instances from different classes are considered irrelevant. 
We run 3 iterations of interaction with 10 feedback samples resulting with 31 few-shot images in the last iteration. As there is randomness in the image selection for labeling, we further run the experiments 5 times with different random seeds and indicate the STD margins. In all our experiments we use 5 different initial queries for each class.
 \begin{figure*}[!ht]
    \centering
    \includegraphics[width=0.22\textwidth]{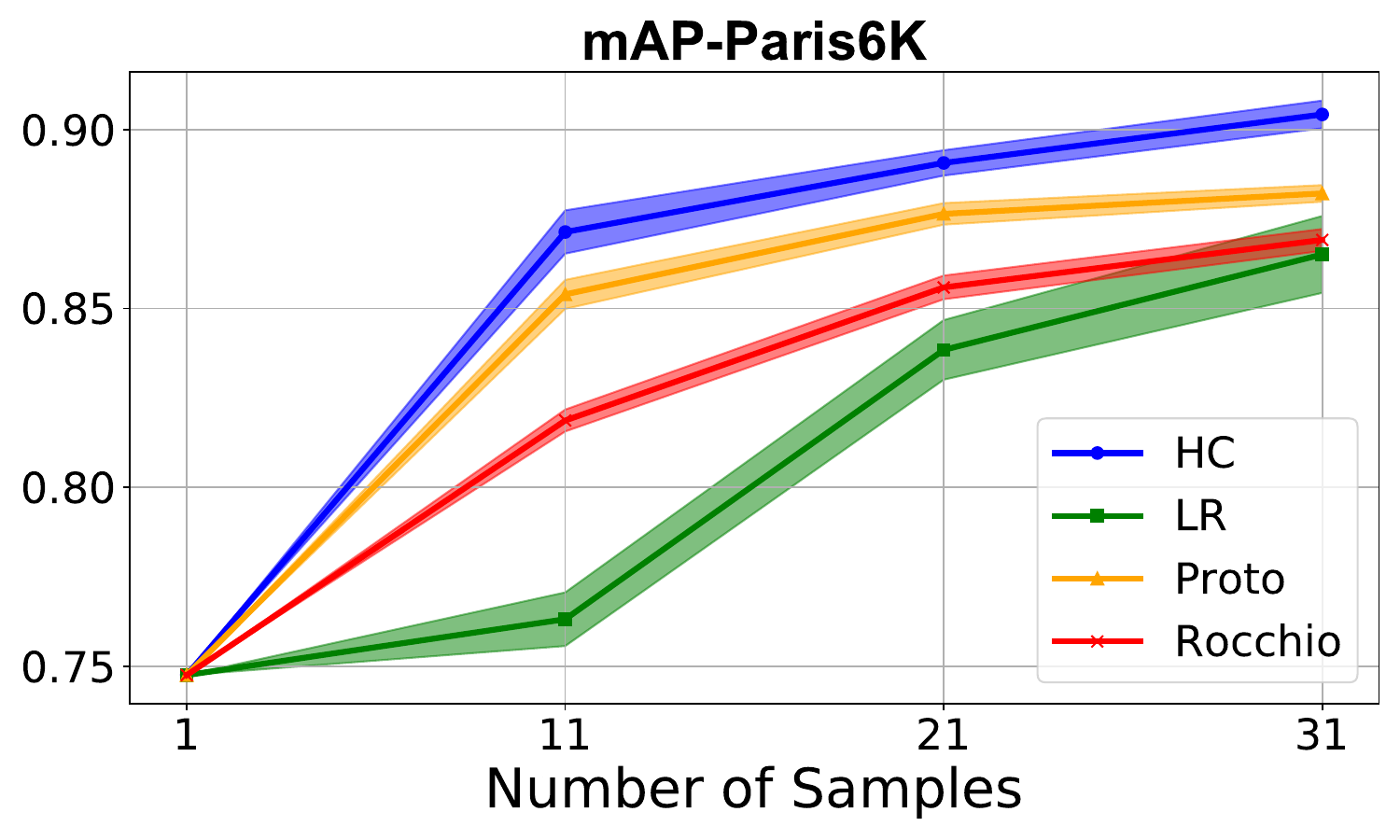}
    \includegraphics[width=0.22\textwidth]{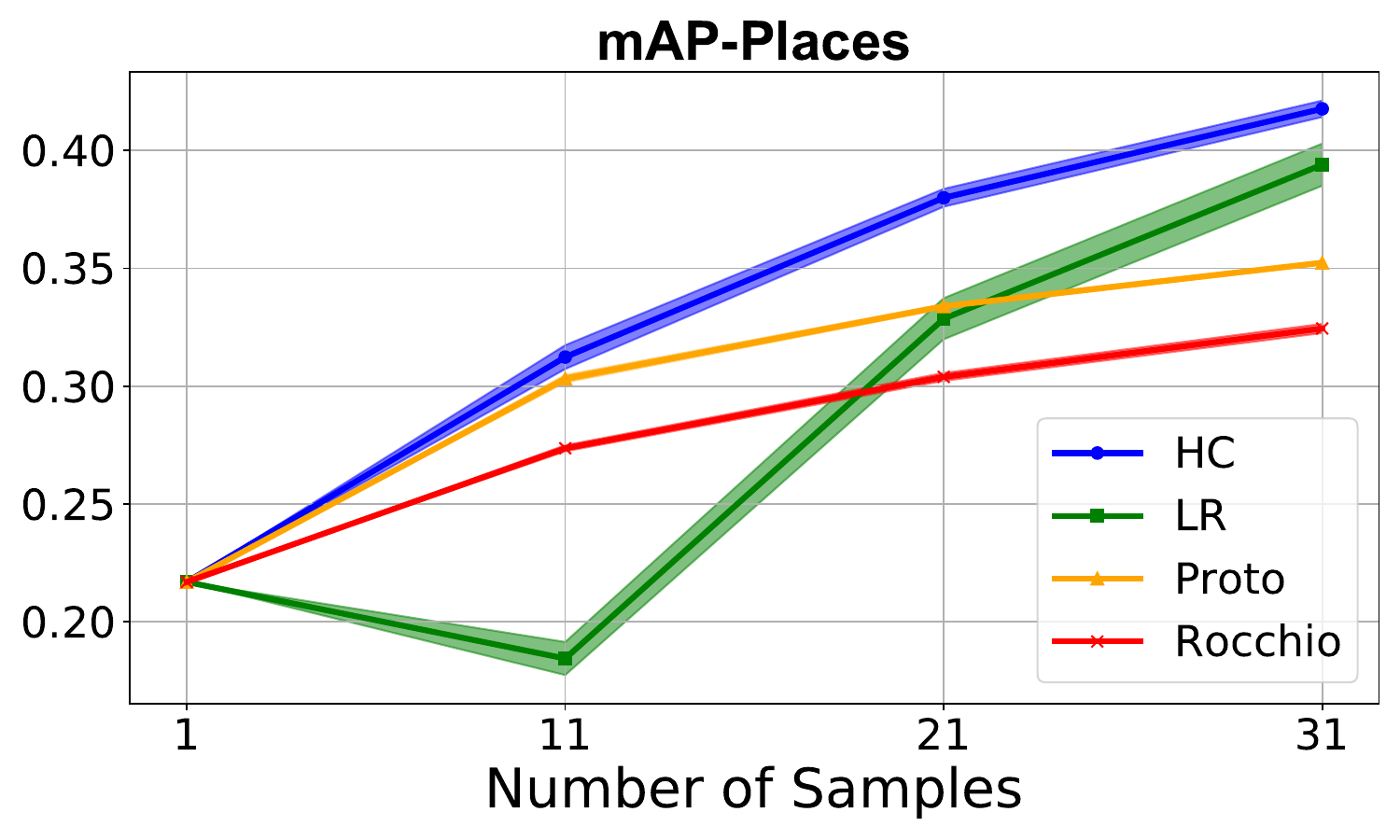}
    \includegraphics[width=0.22\textwidth]{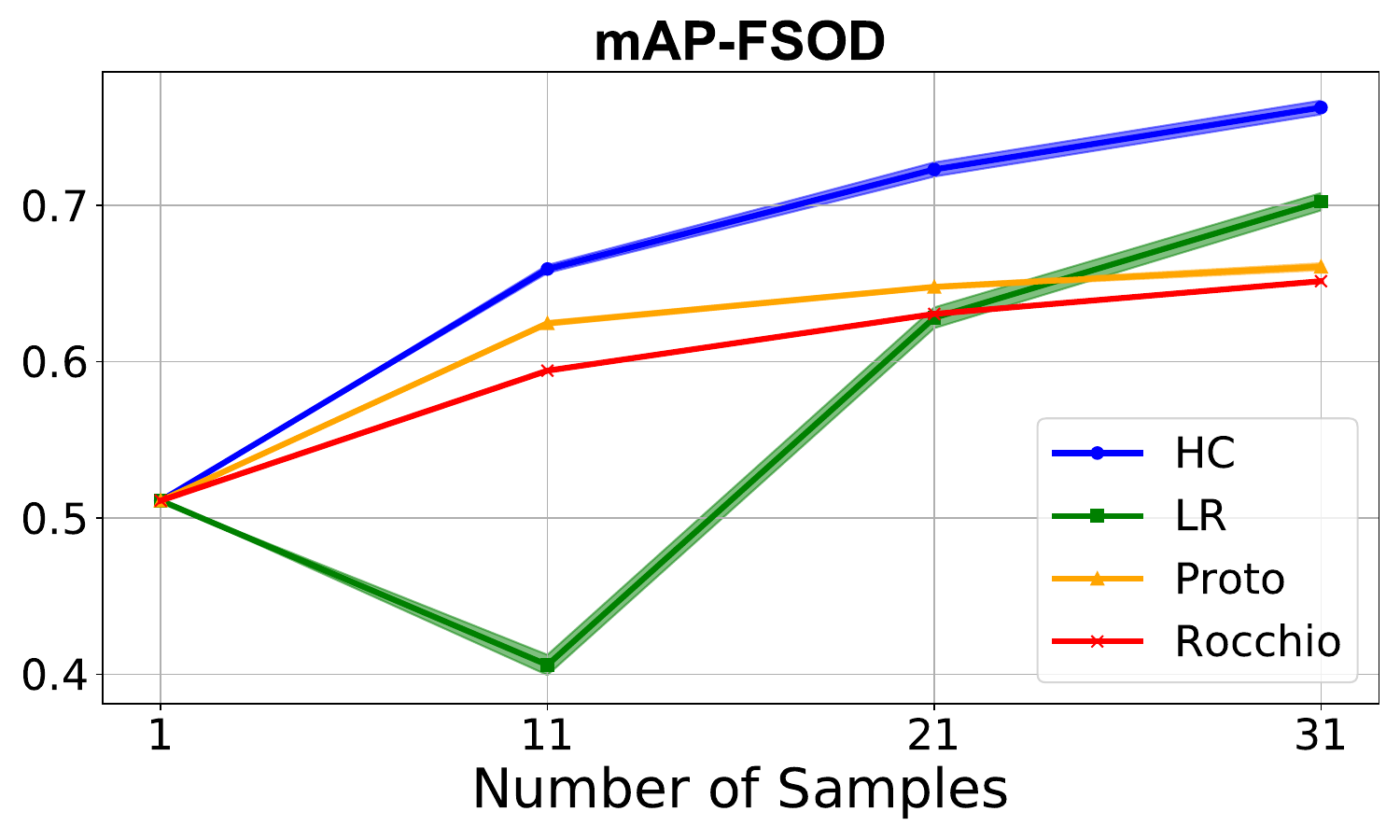}
    \includegraphics[width=0.22\textwidth]{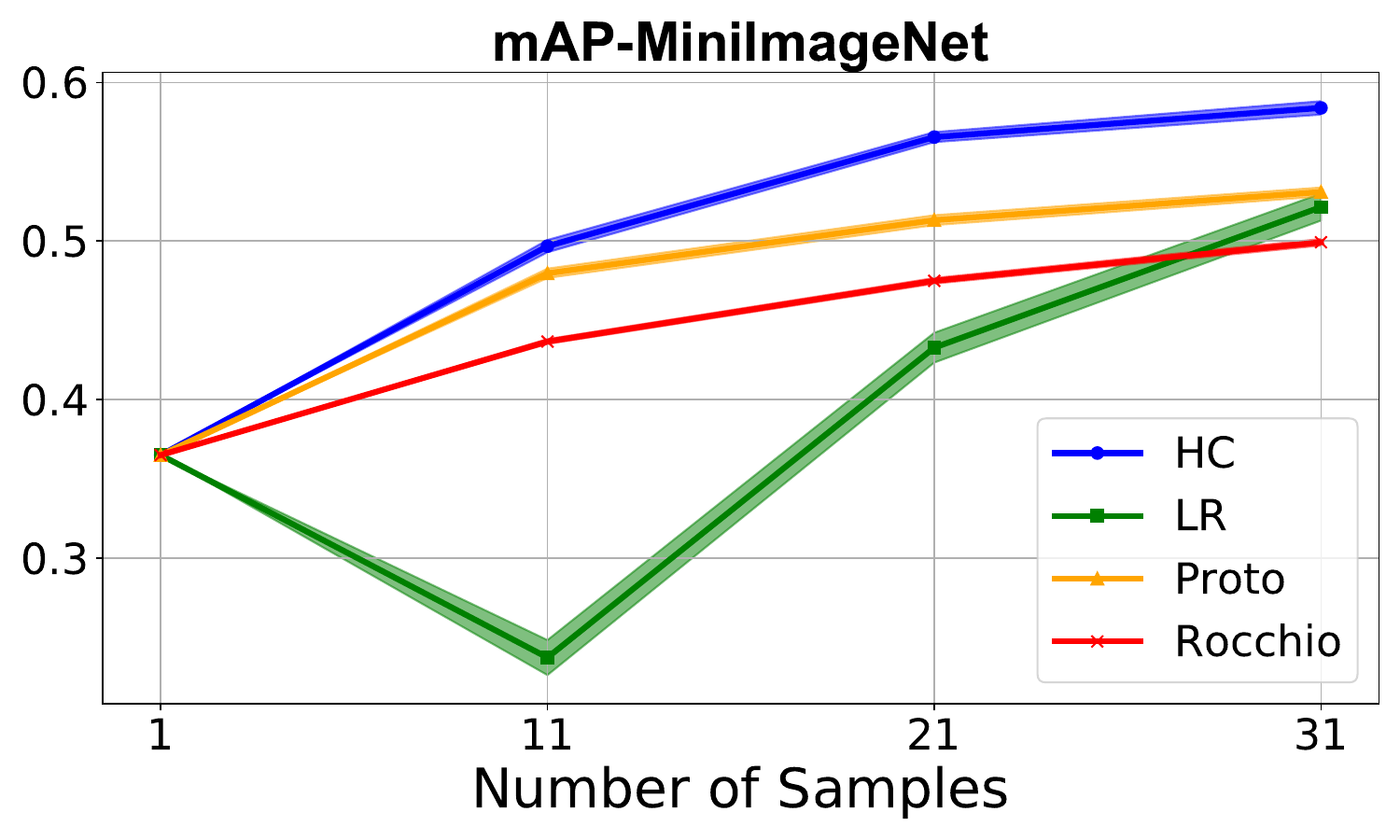} \\
    \includegraphics[width=0.22\textwidth]{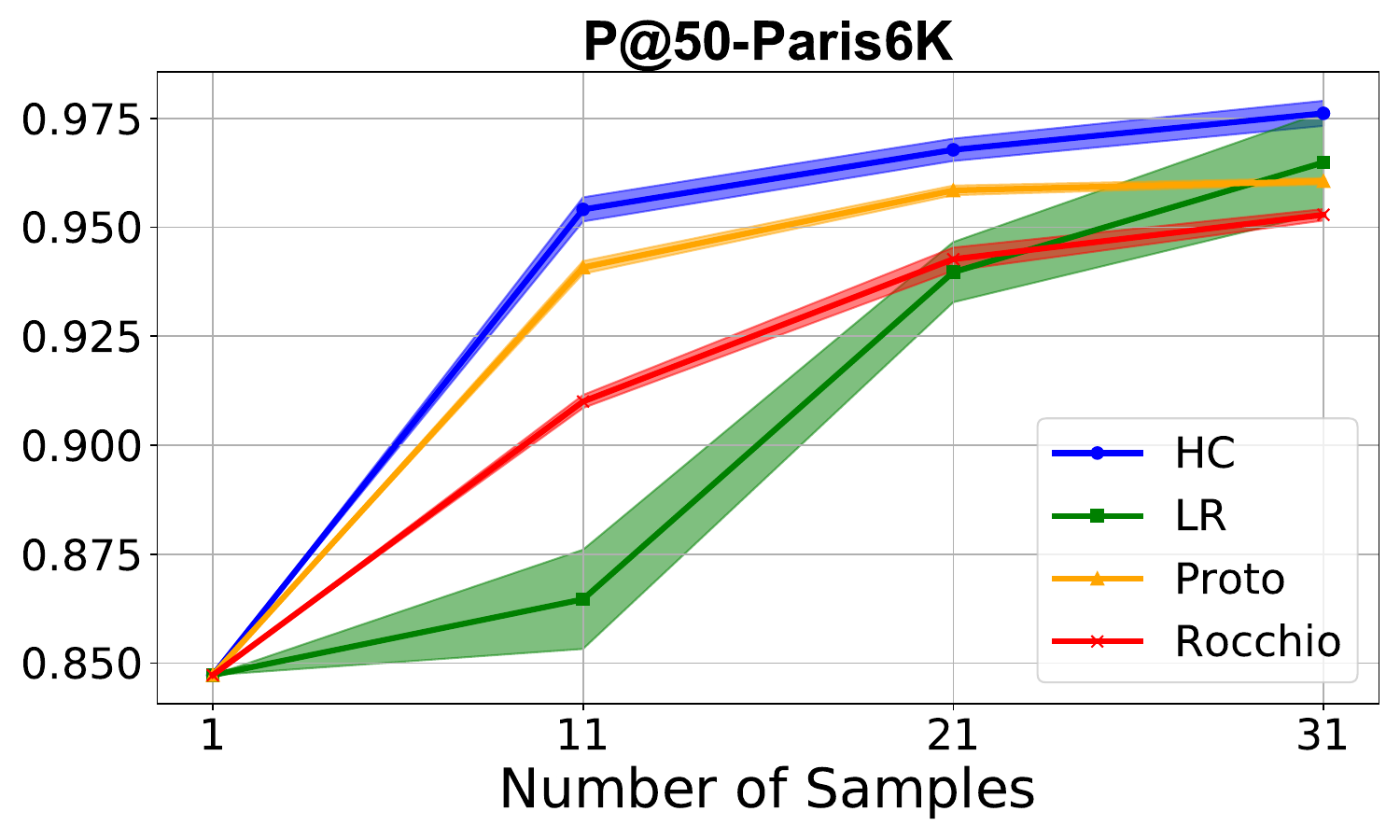}
    \includegraphics[width=0.22\textwidth]{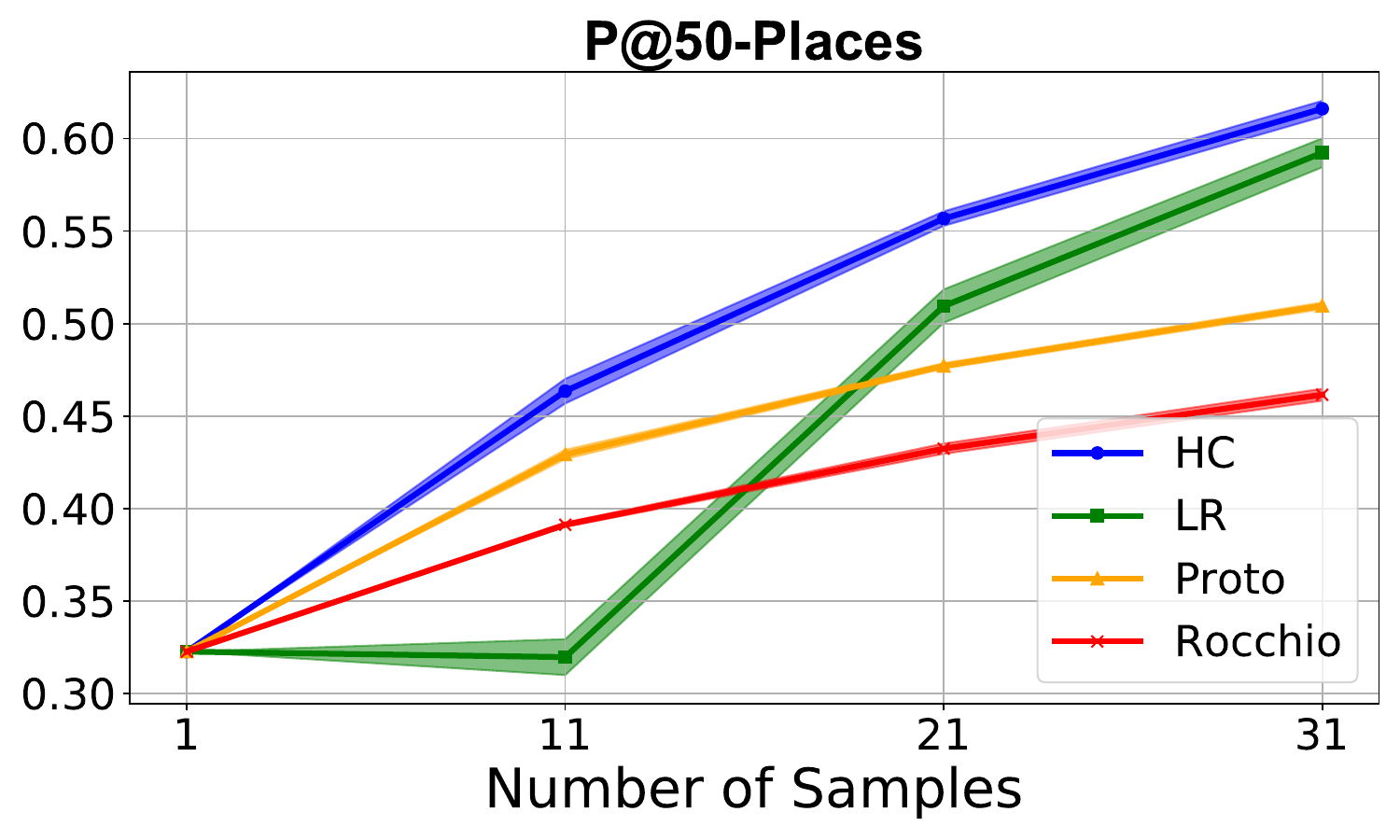}
    \includegraphics[width=0.22\textwidth]{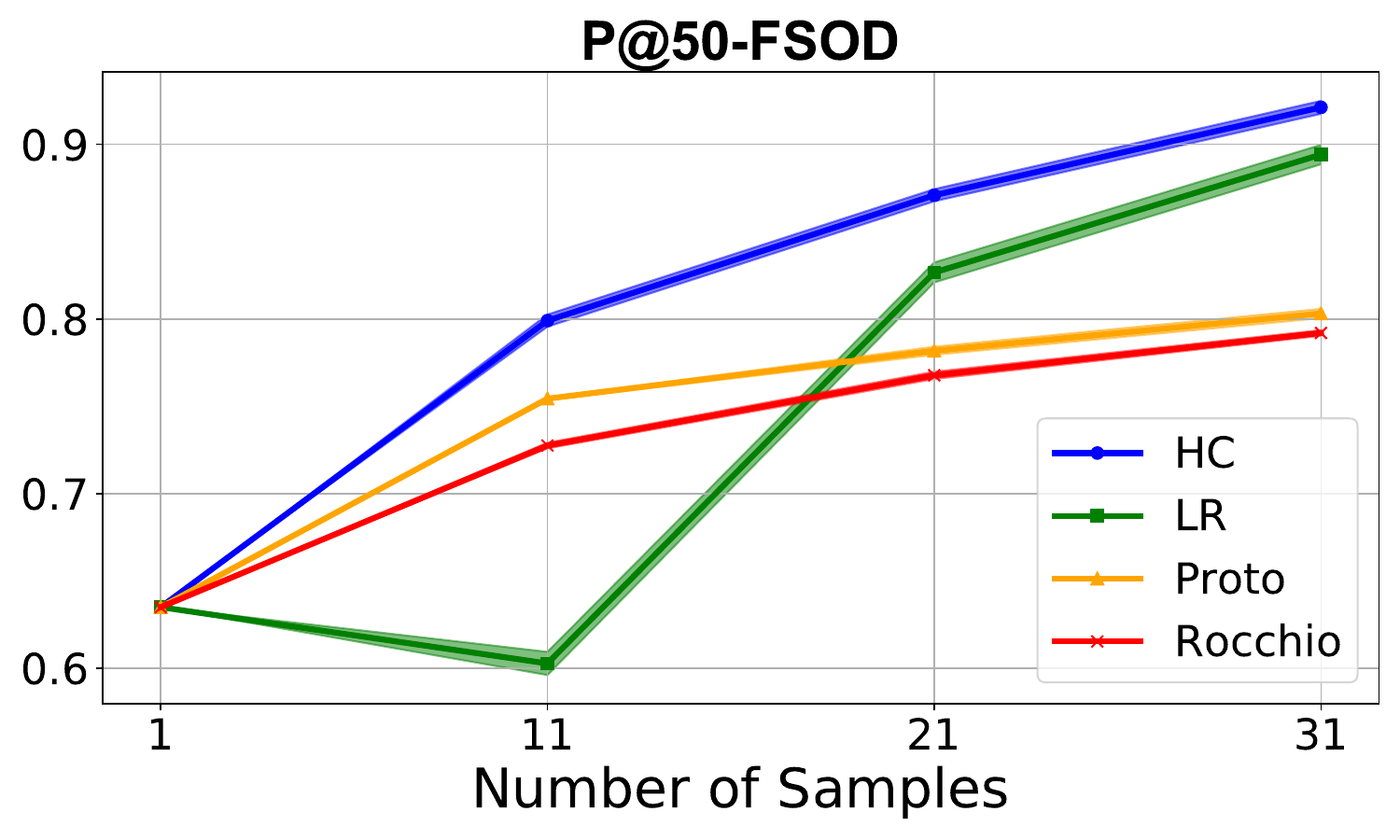}
    \includegraphics[width=0.22\textwidth]{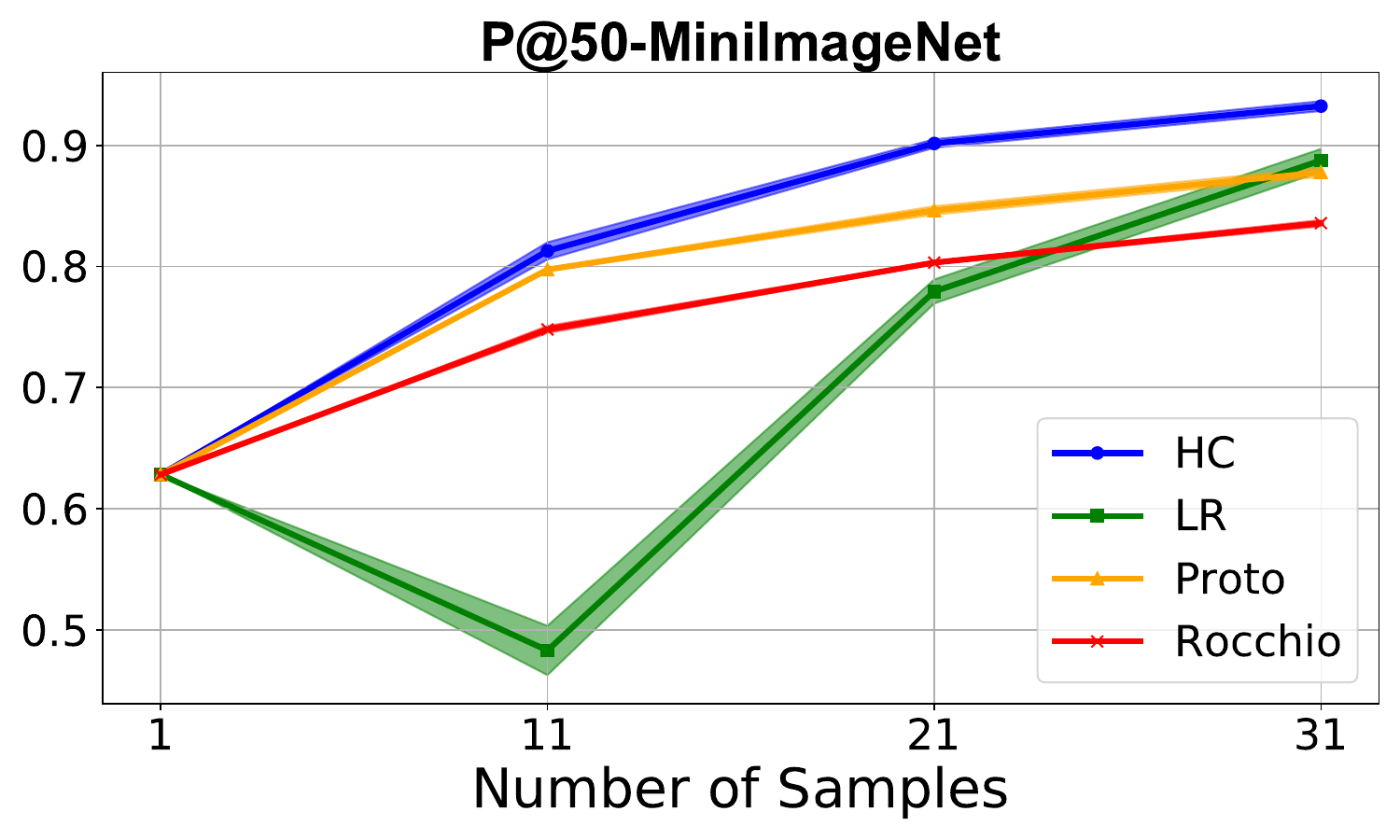}  
    \caption{mAP and Precision@50 on Image Retrieval with Relevance Feedback. We compare our HyperClass to three alternatives. X-axis denotes the number of samples available for training (few-shots). The ribbon, represents the STD for various query seeds. LR and Rocchio were used in prior art,  \cite{putzu2020convolutional} and  \cite{OptimizedQueryIRRF2017} respectively.}
    \label{fig:FSIR}
\end{figure*}

{\bf Compared Methods:} a) {\it IRRF}: Considering lack of standard benchmark and publicly available code for previous IRRF methods, we compare our method to implementation of \cite{putzu2020convolutional} that fine-tunes a linear layer over a pretrained CNN (equivalent to Logisitic Regression), and to the well known Rocchio algorithm used in \cite{OptimizedQueryIRRF2017}.
b) {\it Few-Shot classification baselines}: we opted for Proto, a prototypical approach (defined in Eq. \ref{eq:Proto}) utilizing solely positive samples and DiffSVM \cite{MetaOptNet_CVPR19}, that are computationally cheap for train (adaptation) and inference on a large scale dataset. We train DiffSVM,  based on meta-learning on IRRF episodes similar to HC. In order to maintain consistency in the size of the support set, we choose all the feedback samples in Proto from the positive, potentially offering an advantage due to  increased number of positives.

 
\subsection{Few-Shot One Class Classification}
{\bf Datasets: } For FSOCC, we conducted experiments on two common benchmarks for few-shot learning: MiniImagenet \cite{miniImagenet}, and TieredImageNet \cite{tieredImageNet}, both are subsets of the ImageNet dataset ~\cite{imagenet}. The MiniImagenet dataset consists of 100 categories from ImageNet with 84×84 RGB images, 600 per category. We follow the standard split of training/validation/testing with 64/16/20 classes \cite{ravi2017optimization}. The TieredImageNet dataset has 608 visual categories, split to  351/97/60 training/validation/testing categories, with a total of 450K 84x84 RGB images as in Ye et. al. \cite{tieredImageNet}. For evaluation, we follow the same protocol as in ~\cite{fs_occ-AAAI, fs_occ-kozerawski}, and report the average performance with a 95\% confidence interval on 10K meta-testing tasks, where each task contains a 1-way K-shot support set and a query set with 15 positive and 15 negative samples. We use three evaluation metrics: Accuracy, F1-score, and AUROC (Area Under ROC curve), with thresholds for accuracy and F1-score in our experiments obtained according to the validation set.

{\bf Compared Methods: } Here we compare our method to previous SoTA ~\cite{fs_occ-kozerawski, fs_occ-AAAI} together with two strong baselines that we proposed: One-class SVM (OC-SVM) applied on the pre-trained features, and {\bf Proto}, which takes the mean feature vector of the support set as the classifier (as in IRRF).

\subsection{Few Shot Open Set Recognition}
Few-shot Open set recognition suggests a benchmark that divides the problem into two subtasks, {\it negative detection} (\ie out-of-set classification) and a multi-class closed set classification. For each task there is a separate performance report. The negative detection is a binary classification task separating the ``unseen" out-of-set class samples from the in-set samples belonging to categories in the support set.

Since our HyperClass generates a binary classifier we further implemented a na\"ive generalization to the FSOR negative detection task. We use this benchmark to showcase the capability of our classifier in this unique use case. In contrast to FSOCC, in this scenario the positive (in) set is composed of $N-way$ categories, where commonly $N=5$. We therefore train a different projection head for each support class (with the provided labeled examples). Given a query sample, we then take the max probability over the trained classifiers as in-set probability. Note that we leave out the closed-set recognition task, since these two tasks are separated in the benchmark and any standard few-shot multi-class classification can be used \eg \cite{PMF_CVPR22,GoodEmbedding} to address it. 

{\bf Datasets:} Following FSOR test protocol, we evaluate our method on MiniImageNet \cite{miniImagenet} and TieredImageNet \cite{tieredImageNet} benchmarks with the same train/val/test splits as in FSOCC, and use AUROC measures for the negative detection task. In this scenario, a meta-testing episode is constructed by sampling a regular $N$-way $K$-shot support set, whereas the query set consists of 15 samples from each support class and additional 75 samples from 5 unknown classes (15 each), sampled randomly from the remaining non-support classes. This leads to a total of 75 known queries and 75 unknown queries for a 5-way episode. 

{\bf Compared Methods: } We compare ourselves to a) SoTA methods  \cite{FSOR_cvpr2022,FSOR_cvpr2021,FSOR_cvpr2020} b) Baselines generated from One-Class SVM and Proto. 

\subsection{Implementation Details}
We follow ~\cite{GoodEmbedding} and use the same backbone of ResNet12 ~\cite{resnet} for feature extraction resulting in $d=640$, feature dimension, except for the IRRF realistic protocol, where we use Resnet-50 backbone pre-trained on Imagenet-21K ~\cite{imagenet21k}. 
We use 'learn2learn' package ~\cite{learn2learn} for the implementation of MAML. All of our experiments were run with 1 A100 GPU with 40GB memory on Ubuntu 20.4 machine, each run taking around 3 hours (see suppl. material. for more details).
\subsection{Results}
{\bf IRRF:} We present the results of IRRF in Figure \ref{fig:FSIR}. All methods start with the same single user-query using cosine similarity to rank the database and hence they obtain equal performance. Proto is superior to LR for smaller support sets (initial iterations). However, as more labeled samples are added to the pool, the LR classifier recovers, exploiting also the available negative samples. Nevertheless, HyperClass outperforms all compared methods through all the range. We also tested a comparing method based on DiffSVM \cite{MetaOptNet_CVPR19}. Despite achieving success in standard FSL testing, our evaluation with DiffSVM for IRRF, on MiniImagenet benchmark led to mAP values way below the competitive methods, assumably due to lack of  a prior knowledge learned in the process. We therefore omitted these results from the plot.
The results further show the robustness of our method to cope with in-domain and cross-domain scenarios as well as object-level retrieval. Qualitative examples in Fig. \ref{fig:Visual_and_InferenceScheme}, show the advantage of our method against LR (used in \cite{putzu2020convolutional}). We further show examples of the rank evolution during feedback cycles in suppl. material.

{\bf FSOCC:} Table \ref{tab:FSOCC on miniImagenet} summarizes the results for FSOCC on MiniImagenet and TieredImageNet. 
There are two main observations. 1) Our HyperClass reaches SoTA performance with much shallower (vanilla) Conv-4 backbone (64 features), used also in \cite{GoodEmbedding}
2) Employing a deeper backbone, our simple baselines already surpass the previous proposed methods, with our method further boosting the performance on both datasets.
Our transductive HyperClass variant allows a further performance gain, with a stronger impact on 1-shot. 
Finally, we add to the table an upper bound to this task, obtained by training a one-class SVM with 500 positive samples in the support set, instead of 1/5. Interestingly, these results are not far from our results. 
\begin{table*}[!ht]
\begin{center}
\scriptsize
\begin{tabular}{|l|c|c|c|c|c|c|c|}
\hline
 & & \multicolumn{3}{c|}{1-shot} & \multicolumn{3}{c|}{5-shot} \\
\hline
\multicolumn{8}{|c|}{MiniImageNet} \\
\hline
Model & Backbone & Acc & F1-Score  & AUROC & Acc & F1-Score  & AUROC \\ \hline
Upper Bound & ResNet12 & 86.15 $\pm$ 0.1 &  88.02 $\pm$ 0.16 & 94.45 $\pm$ 0.22 & 86.15 $\pm$ 0.1 &  88.02 $\pm$ 0.16 & 94.45 $\pm$ 0.22 \\
\hline
Meta BCE $^1$ & Conv-64  & 57.57 $\pm$ 0.78 &  28.3 $\pm$ 1.9 & 76.4 $\pm$ 1.6 & 77.38 $\pm$ 0.67 & 79.3 $\pm$ 0.5 & 85.0 $\pm$ 0.7 \\
OCML$^1$ & Conv-64 & 68.05 $\pm$ 0.99 & 62.0 $\pm$ 1.4 & 75.9 $\pm$ 1.2 & 74.74 $\pm$ 0.88 & 69.3 $\pm$ 1.3 & 85.4 $\pm$ 0.9 \\
OC-MAML$^2$  & Conv-4$^*$ & 69.1 & - & - & 76.2 & - & - \\ 
\hline
OC-SVM & Conv-4 & 50.12 $\pm$ 0.02 & 66.72 $\pm$ 0.01 & 77.44 $\pm$ 0.29 & 75.16 $\pm$ 0.21 & 76.79 $\pm$ 0.20 & 84.68 $\pm$ 0.20 \\
ProtoNet & Conv-4 & 66.73 $\pm$0.22 & 71.27$\pm$0.24 & 77.48$\pm$0.29 & 69.89 $\pm$ 0.18 & 75.23 $\pm$ 0.17 & 84.48 $\pm$ 0.21 \\
HC (Ours) & Conv-4 & 68.03 $\pm$ 0.24 & 69.05 $\pm$ 0.33 & 78.45 $\pm$ 0.29 & 77.69 $\pm$ 0.19 & 76.84 $\pm$ 0.23 & 86.83 $\pm$ 0.18 \\ \hline
OC-SVM & ResNet12 & 75.14 $\pm$ 0.24 & 72.74 $\pm$ 0.35 & 85.23 $\pm$ 0.23 & 81.36 $\pm$ 0.19 & 82.59 $\pm$ 0.17 & 92.10 $\pm$ 0.12 \\
ProtoNet & ResNet12 & 76.12 $\pm$ 0.24 & 72.02 $\pm$ 0.38 & 85.23 $\pm$ 0.23 & 82.20 $\pm$ 0.19 & 79.78 $\pm$ 0.28 & 92.11 $\pm$ 0.12 \\
HC (Ours) & ResNet12 & {\bf 76.66 $\pm$ 0.23} & {\bf 76.22 $\pm$ 0.31} & {\bf 85.72 $\pm$ 0.23} & {\bf 84.93 $\pm$ 0.15} & {\bf 85.48 $\pm$ 0.15} & {\bf 92.98 $\pm$ 0.11} \\
HC - Transuctive (Ours) & ResNet12 & \textcolor{blue}{{\bf 78.74 $\pm$ 0.24}} & \textcolor{blue}{{\bf 77.95 $\pm$ 0.23}} & \textcolor{blue}{{\bf 89.59 $\pm$ 0.15}} & \textcolor{blue}{{\bf 85.42 $\pm$ 0.18}} & \textcolor{blue}{{\bf 86.12 $\pm$ 0.13}} & \textcolor{blue}{{\bf 93.23 $\pm$ 0.11}} \\
\hline
\multicolumn{8}{|c|}{tieredImageNet} \\
\hline
Upper Bound & ResNet12 & 86.33 $\pm$ 0.17 &  88.67 $\pm$ 0.14 & 95.49 $\pm$ 0.27 & 86.33 $\pm$ 0.17 &  88.67 $\pm$ 0.14 & 95.49 $\pm$ 0.27 \\
\hline
Meta BCE $^1$ & Conv-64 & 55.87 $\pm$ 0.38 &  20.8 $\pm$ 1.0 & 75.5 $\pm$ 0.9 & 75.62 $\pm$ 0.65 &  77.4 $\pm$ 0.5 & 83.1 $\pm$ 0.7 \\
OCML $^1$ & Conv-64  & 72.13 $\pm$ 0.33 & 72.4 $\pm$ 0.3 & 80.0 $\pm$ 0.4 & 78.89 $\pm$ 0.67 & 80.3 $\pm$ 0.6 & 87.8 $\pm$ 0.7 \\ \hline
OC-SVM & Conv-4 & 68.66 $\pm$ 0.24 & 72.18 $\pm$ 0.25 & 79.63 $\pm$ 0.28 & 77.02 $\pm$ 0.19 & 74.22 $\pm$ 0.25 & 87.17 $\pm$ 0.18 \\
ProtoNet & Conv-4 & 71.13 $\pm$ 0.24 & 71.95 $\pm$ 0.31 & 79.66 $\pm$ 0.28 & 77.66 $\pm$ 0.19 & 79.08 $\pm$ 0.21 & 87.10 $\pm$ 0.18 \\
HC (Ours) & Conv-4 &  70.98 $\pm$ 0.25 & 70.76 $\pm$ 0.34 & 80.48 $\pm$ 0.28 & 79.53 $\pm$ 0.19 & 81.28 $\pm$ 0.17 & 88.62 $\pm$ 0.17 \\ \hline
OC-SVM & ResNet12 & 78.39 $\pm$ 0.25 & 75.52 $\pm$ 0.38 & 87.94 $\pm$ 0.22 & 84.50 $\pm$ 0.17 & 84.70 $\pm$ 0.18 & 93.59 $\pm$ 0.12 \\
ProtoNet & ResNet12 & 76.29 $\pm$ 0.26 & 68.58 $\pm$ 0.46 & 87.95 $\pm$ 0.22 & 80.24 $\pm$ 0.25 & 73.45 $\pm$ 0.45 & 93.49 $\pm$ 0.13 \\
HC (Ours) & ResNet12 & {\bf 79.01 $\pm$ 0.24} & {\bf 76.66 $\pm$ 0.36} & {\bf 88.40 $\pm$ 0.22} & {\bf 84.77 $\pm$ 0.18} & {\bf 86.58 $\pm$ 0.14} & {\bf 94.11 $\pm$ 0.11} \\
HC - Transductive (Ours) & ResNet12 & \textcolor{blue}{{\bf 80.07 $\pm$ 0.22}} & \textcolor{blue}{{\bf 78.93 $\pm$ 0.25}} & \textcolor{blue}{{\bf 91.16 $\pm$ 0.15}} & \textcolor{blue}{{\bf 84.93 $\pm$ 0.19}} & \textcolor{blue}{{\bf 87.12 $\pm$ 0.12}} & \textcolor{blue}{{\bf 94.32 $\pm$ 0.13}} \\
\hline
\end{tabular}
\caption{Few-Shot One-Class Classification results on {\bf MiniImagenet} and {\bf tieredImagenet} for various methods and backbones. $^{[1]}$ \cite{fs_occ-kozerawski}, $^{[2]}$ \cite{fs_occ-AAAI}. Performance measures correspond to confidence interval of 95\% over 10K episodes. OC-MAML reports only accuracy measure. Note that Conv-64 and Conv-4* are larger/optimized architectures w.r.t our simple Conv-4, commonly used in MAML methods. We also report the results with our transductive version marked in blue and an upper bound is computed by training OC-SVM on 500 support samples, thus its results are repeated for 1/5 shot. }
\label{tab:FSOCC on miniImagenet}
\end{center}
\end{table*}

{\bf FSOR:} Table \ref{tab:FSOR} shows the AUROC for negative detection task in FSOR. We reach a performance comparable to SoTA on MiniImageNet and surpass previous work on TieredImageNet in 5-shot (with 82.83\% vs. 81.64\%) and are comparable to SoTA on 1-shot (with 74.54\% vs. 74.95\%). HyperClass also obtains lower STD values, indicating a higher statistical validity of our results.  

Interestingly, our conducted simple baselines of OC-SVM and Proto show very competitive results, a fact that to the best of our knowledge has been overlooked so far. We omit SEMAN-G results in ~\cite{FSOR_cvpr2022} from our table for fair comparison, as it makes use of class names (with a pre-trained language model), an information that is not supplied by the standard protocol.
\begin{table}[!ht]
\begin{center}
\scriptsize
\begin{tabular}{|l|c|c|c|c|}
\hline
& \multicolumn{2}{c|}{MiniImageNet 5-way} & \multicolumn{2}{c|}{TieredImageNet 5-way} \\ \hline
Model & 1-shot & 5-shot & 1-shot & 5-shot \\ \hline
PEELER $^1$  & 60.57$\pm$0.83 & 67.35 & 65.20 & 73.27 \\
SnaTCHer-F $^2$  & 68.27$\pm$0.96 & 77.42 & 74.28 & 82.02 \\
SnaTCHer-T $^2$ & 70.17$\pm$0.88 & 76.66 & 74.84 & \underline{82.03} \\
SnaTCHer-L $^2$ & 69.40$\pm$0.92 & 76.15 & \bf{74.95} & 80.81 \\
ATT $^3$ & \underline{71.35$\pm$0.68} & \underline{79.85$\pm$0.58} & 72.74$\pm$0.78 & 78.66$\pm$0.65 \\
ATT-G $^3$ & {\bf 72.41$\pm$0.72} & {\bf 79.85$\pm$0.57} & 73.43$\pm$0.78 & 81.64$\pm$0.63 \\ \hline
OC-SVM   & 69.26$\pm$0.68 & 72.58$\pm$0.59 & 71.58$\pm$0.57 & 75.00$\pm$0.49 \\
Proto  & 70.59$\pm$0.64 & 76.74$\pm$0.55 & \underline{74.85$\pm$0.53} & 81.40$\pm$0.42 \\
HC (Ours)   & 70.29$\pm$0.26 & 79.21$\pm$0.50 & 74.54$\pm$0.54 & \bf{82.83$\pm$0.38} \\
\hline
\end{tabular}
\caption{AUROC for FSOR in-set classification subtask (\%) with confidence interval of 95\% over 600 episodes. Top performance are in bold, and 2nd place are underlined. $^{[1]}$ ~\cite{FSOR_cvpr2020}, $^{[2]}$ \cite{FSOR_cvpr2021}, $^{[3]} $\cite{FSOR_cvpr2022}}.  
\label{tab:FSOR}
\end{center}
\end{table}
\begin{figure}[!ht]
    \centering
    \includegraphics[width=0.3\textwidth]{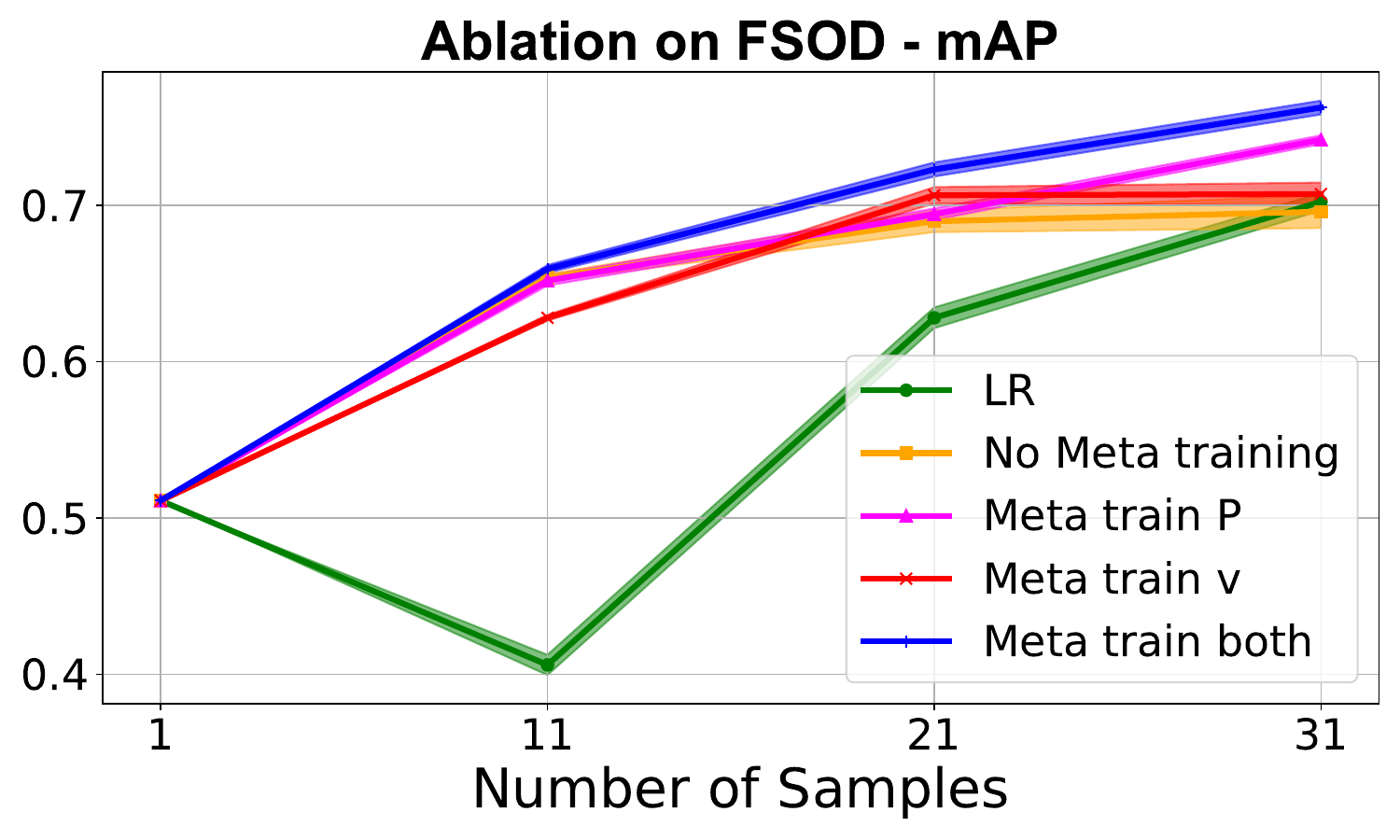}
    \caption{Ablation study on different training strategies of HyperClass tested on IRRF. The colored ribbons indicate the STD from various query seeds.}
    \label{fig:FSIR_Ablation}
\end{figure}
\subsection{Ablation}
The two main components of our method are the global classifier $v$ and the projection head $P$, with our aim to initialize them effectively for quick adaptation on novel tasks. We wish to quantify the benefit of our training strategy leveraging meta-training over tasks.
To this end, we conduct several experiments on FSOD dataset. Figure \ref{fig:FSIR_Ablation} shows the IRRF performance, where we consider four variants. The first one is without any meta-training, initializing $P$ and $v$ randomly in meta-testing. The two other variants denoted "Meta-train $v$" and "Meta-train $P$" refer to the initialization of either $v$ \textbf{or} $P$ through meta-training. Finally, "Meta-train both" is the complete method, where we learn an initialization of both $v$ and $P$ in the meta-training stage. Note that all variants use the same optimization (adaptation) in meta-testing stage, namely a few updates for $P$. Figure \ref{fig:FSIR_Ablation} shows that without meta-training (orange line), the performance converges to the same level as LR's in the last iteration, while still adding value in the first iterations. This shows the advantage of the classifier decomposition to $v$ and $P$ even without meta learning. Meta training only $v$ or $P$ improves the results (red and purple lines), with $P$ being more significant. Finally, meta training both $v$ and $P$ (blue line) yields the best results, with highest mAP and Precision@50. This clearly emphasizes the advantage of 'learning to learn' strategy here, where both the task-agnostic classifier and the adaptive projection head leverage the information gained globally over tasks.  

\section{Summary and Discussion}
Image retrieval with Relevance Feedback (IRRF) aims to tune the search results according to user's feedback. IRRF contains a unique combination of challenges for modern machine learning methods, including learning from few labeled samples, in an imbalanced and open-set setting. In this study we addressed the classifier refinement toward a given query concept, based on few examples.
 Drawing inspiration from the domain of few-shot learning, we suggested a novel approach for IRRF. Our method  incorporates the training of a global, task-agnostic classifier alongside a quick adaptation model designed to fine-tune the classifier from the accumulated feedback samples. We evaluate our model on IRRF task, outperforming strong baselines. We further show the versatility of our approach, easily adapting it to few-shot one-class classification and open-set recognition. 

 Limitations of our method may include the specificity of the method to binary classification (not multi-class) and since our model is global-feature based and not detector-based, it may miss small objects (recall).
 
 Being based on a fixed and pre-trained backbone, we also believe that our method can be extended to other modalities and applications \eg audio and video. 
\bibliography{arxiv_version}
\include{arxiv_suppl}
\end{document}

%% file: arxiv_suppl.tex
\renewcommand{\figurename}{Fig. S}
\renewcommand\tablename{Table S}
\newcommand{\figref}[1]{S\ref{#1}}
\newcommand{\tabref}[1]{S\ref{#1}}
\newcommand{\cmark}{\ding{51}}%
\newcommand{\xmark}{\ding{55}}

\appendix
\numberwithin{equation}{section}
{\huge{\textbf{Supplemental Material}}}
\newline
\newline

In this supplementary material we give a background on few-shot learning (Sec. A). Next, in Sec. B, we present the derivation in our Theoretical Analysis in the main paper (Sec. 3.1 in the paper). We further show in Sec. C results on IRRF with different ratios of positive feedback samples. Sec. D in this suppl. material shows examples of rank evolution along the feedback cycles. Finally, we conclude in Sec. E by providing implementation details.
\section{Background and preliminaries on Few Shot Learning}
Few-Shot Learning (FSL) is a framework that enables a pre-trained model to generalize over new categories of data (that the pre-trained model has not seen during training) using only a few labeled samples per class.  Few shot learning methods are typically applied and tested in a closed set form where the task is classification between $N$ novel unseen classes (often five), given few samples ($K$ shots) from each class. The most common evaluation for few-shot image classification is called {\it meta-testing}, where one is evaluated on randomly sampled $N$-way classes and $K$-shot sample per-class of novel classes. The $N$-way $K$-shot samples are called {\it support set} and are used for adaptation, whereas different samples from those $N$ classes are used for evaluation, and are part of the {\it query set}. Note that this terminology is different from user query in image retrieval task. For user query we will use the explicit term. 

Few-shot learning has been actively investigated for classification problems and can be categorized into two main approaches, Meta Learning and Transfer Learning. 
In Meta Learning, the idea is to mimic the testing procedure during training ~\cite{fs_transformer, differential_svm, MAML, proto_net} by episodic sampling of $N$-way $K$-shot tasks from the base classes (usually a large labeled set).  In the IRRF task the relevance feedback samples at certain search iteration can be regarded as the support set, while the samples in the corpus introduce the query set.

Meta Learning is further divided into two lines of solutions, one based on metric learning and the other attempts to "learn how to learn".
In Metric Learning, one seeks to optimize the embedding space to be generalizable to new classes with evaluation based on simple classifiers \eg Prototypical Networks ~\cite{proto_net} or differentiable SVM ~\cite{differential_svm}. On the other hand, Model-Agnostic Meta-Learning (MAML) ~\cite{MAML} suggests an optimization procedure aiming to learn an initialization that can be quickly adapted to a new task. They use two loops to conduct the training, known as the {\it inner} and {\it outer} loops. During the inner loop, weight updates are performed on the support set per task. At the outer loop, the weights are updated by the mean gradient of the loss (across the meta-batch of tasks) over the query set, based on gradients accumulated during inner loops. Yet, MAML is hard to train, with common limitation of shallow backbone \eg Conv-4, compromising features quality ~\cite{MAML++, fs_occ-kozerawski}. Our method makes use of MAML strategy in the feature space for fast adaption to a new user-labeled image set for retrieval. 

While many initial FSL works follow meta-learning frameworks, recent studies suggest that the standard transfer learning is a strong baseline for few-shot classification \cite{GoodEmbedding,simpleFSL_ICCV21}. Such transfer learning methods pre-train a model using all available training classes and leverage the model for testing. The model is then adapted to a new task often by training a linear classifier as suggested in ~\cite{GoodEmbedding, fs_baseline, chen2019closer, simpleFSL_ICCV21}. Recent methods suggest a combination of both \cite{PMF_CVPR22,MetaBaseline2021}, Pretrain, Meta-Learn and Finetune in few shot image classification. We follow this strategy, however our scheme divides the learning pipeline into two separate and mutually exclusive stages, first building a good embedding and then learning an adaptation module that is specialized for the fixed embedding space.

In this study we suggested a novel method for IRRF inspired by the techniques in the field of few-shot learning (FSL). In our task, a small training set is constructed from the feedback cycles with the user, in which a few samples are labeled. 

{\bf Applicability of few-shot classification methods to IRRF task:  } 
The majority of FSL methods are based on metric learning. 
The key idea of FSL metric-based methods \eg \cite{proto_net, 1way_proto,EPFNET_ECCV20} is to learn an embedding that can separate classes in general. Nevertheless, metric learning techniques frequently employed in FSL rely on unimodal classes which is not the case for the negative set in the IRRF task. However, a prototypical approach can be employed when using solely positive samples. Thus, we utilize this approach to construct a baseline for the IRRF task. 
Some FSL methods are based on adaptation/training a classifier on the support set without online adaptation of the embeddings \cite{DeepEMD_CVPR20,MetaOptNet_CVPR19, GoodEmbedding}. As these methods tend to better comply with the IRRF constraints, we create a baseline based on this type of FSL methods \cite{differential_svm} in our experiments.
\section{Theoretical Analysis}
In this section we present the derivation of the classifier update schemes, given in equations (6,8,9) in the main paper. Then, we show the intuition behind our method.

\subsection{Loss Derivation}
We start with derivation of the loss for a simple linear (and binary) classifier represented by $W \in \mathbb{R}^d$, with $d$ denoting the feature dimension. Let us further denote the classifier predicted probability by $\sigma(z)$, where $z:=W^T x$ and $\sigma(\cdot)$ is a sigmoid function:
\begin{equation}
    \label{eq:Sigmoid}
    \sigma(z) = \frac{e^z}{e^z+1}.
\end{equation}
The Binary Cross Entropy (BCE) loss for a single {\it positive} sample in the support set $x \in x^s_p$ is :
\begin{eqnarray}
\label{eq:Loss}
\mathcal{L} = -\log{\frac{e^z}{e^z+1}} = -z + \log({e^z+1}).
\end{eqnarray}

Now we can derivate $\mathcal{L}$ w.r.t the parameters $W$:
\begin{equation}
\nabla_W \mathcal{L} = -x + \frac{e^z}{e^z+1} x.
\end{equation}

Similarly, for a {\it negative} sample $x \in x^s_n$, we obtain:
\begin{align}
    &\mathcal{L} = -\log{\frac{1}{e^z+1}} = \log({e^z+1}) \\ 
    &\nabla_W \mathcal{L} = \frac{e^z}{e^z+1} x.
\end{align}

In order to combine positive and negative samples in a single compact notation, we define:
\begin{equation}
\lambda_x := \begin{cases} \alpha ,& \text{if  } x \in x_p^s \\ \alpha-1 ,& \text{if  } x \in x_n^s \end{cases}
\end{equation}
where we use the following definition in the paper, $\alpha:=1-\sigma(z)$.
Then, the gradient for any $x \in x^s$ becomes:
\begin{equation}
\nabla_W\mathcal{L} = -\lambda_x \, x.
\end{equation}
The update scheme for $W$ at gradient step $t+1$ (assuming unit learning rate, without loss of generality), over all the support samples (both classes), $x^s = x_s^p \bigcup x^s_n$ is then: 
\begin{align}
\Delta W &= W_{t+1} - W_t  = \frac{1}{|x^s|}\sum_{x\in x^s}\lambda_x \, x \\ \nonumber
                          &= \frac{1}{|x^s|}\sum_{x\in x^s_p} \alpha x - \frac{1}{|x^s|}\sum_{x\in x^s_n} (1-\alpha)x.
\end{align}
This expression represents the update scheme for Logistic Regression as we describe in Eq. (6) in the paper.

Now let us consider our {\it HyperClass} model where:
\begin{equation}
    z = (Pv+b)^T x
\end{equation}

Substituting $z$ in \eqref{eq:Loss} and derivating according to $P, v, b$ for a support sample yields:
\begin{align}
    &\nabla_{v}\mathcal{L} = -\lambda_x P^Tx \\ \nonumber
    &\nabla_{P}\mathcal{L} = -\lambda_x xv^T \\ \nonumber
    &\nabla_{b}\mathcal{L} = -\lambda_x x,
\end{align}
and summing over all the support samples yields:
\begin{align}
\label{eq:GradientsPvb}
&\nabla_{v}\mathcal{L} = \frac{1}{|x^s|}\sum_{x\in x^s} -\lambda_x P^Tx \\ \nonumber
&\nabla_{P}\mathcal{L} = \frac{1}{|x^s|}\sum_{x\in x^s}-\lambda_x xv^T \\ \nonumber
&\nabla_{b}\mathcal{L} = \frac{1}{|x^s|}\sum_{x\in x^s}-\lambda_x x.
\end{align}

\begin{figure*}[!ht]
    \begin{subfigure}[b]{0.45\textwidth}
    \includegraphics[width=\textwidth]{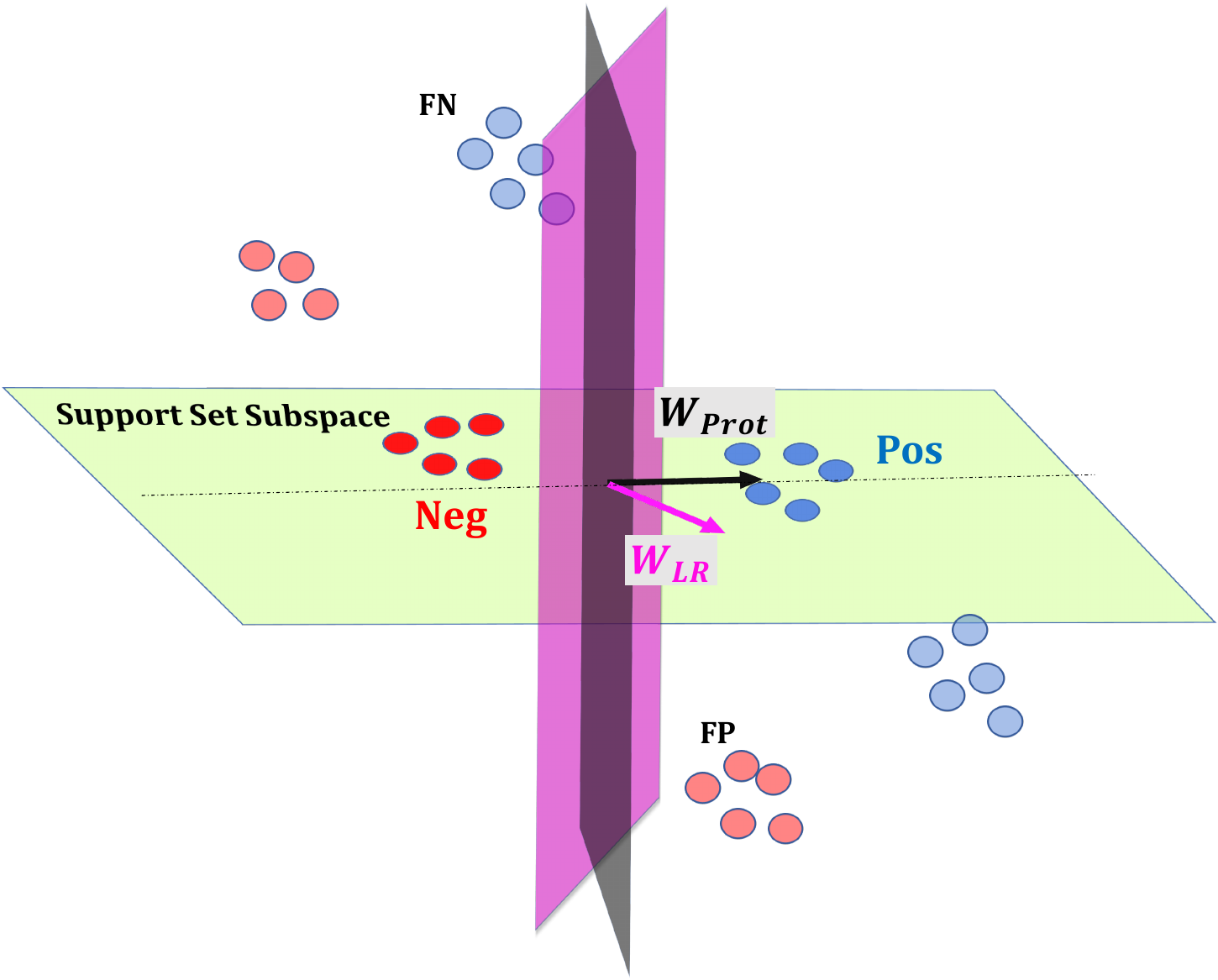}
    \caption{Linear Classifier}
    \label{fig:LinearClass_Illustration}
    \end{subfigure}
    \begin{subfigure}[b]{0.45\textwidth}
    \includegraphics[width=\textwidth]{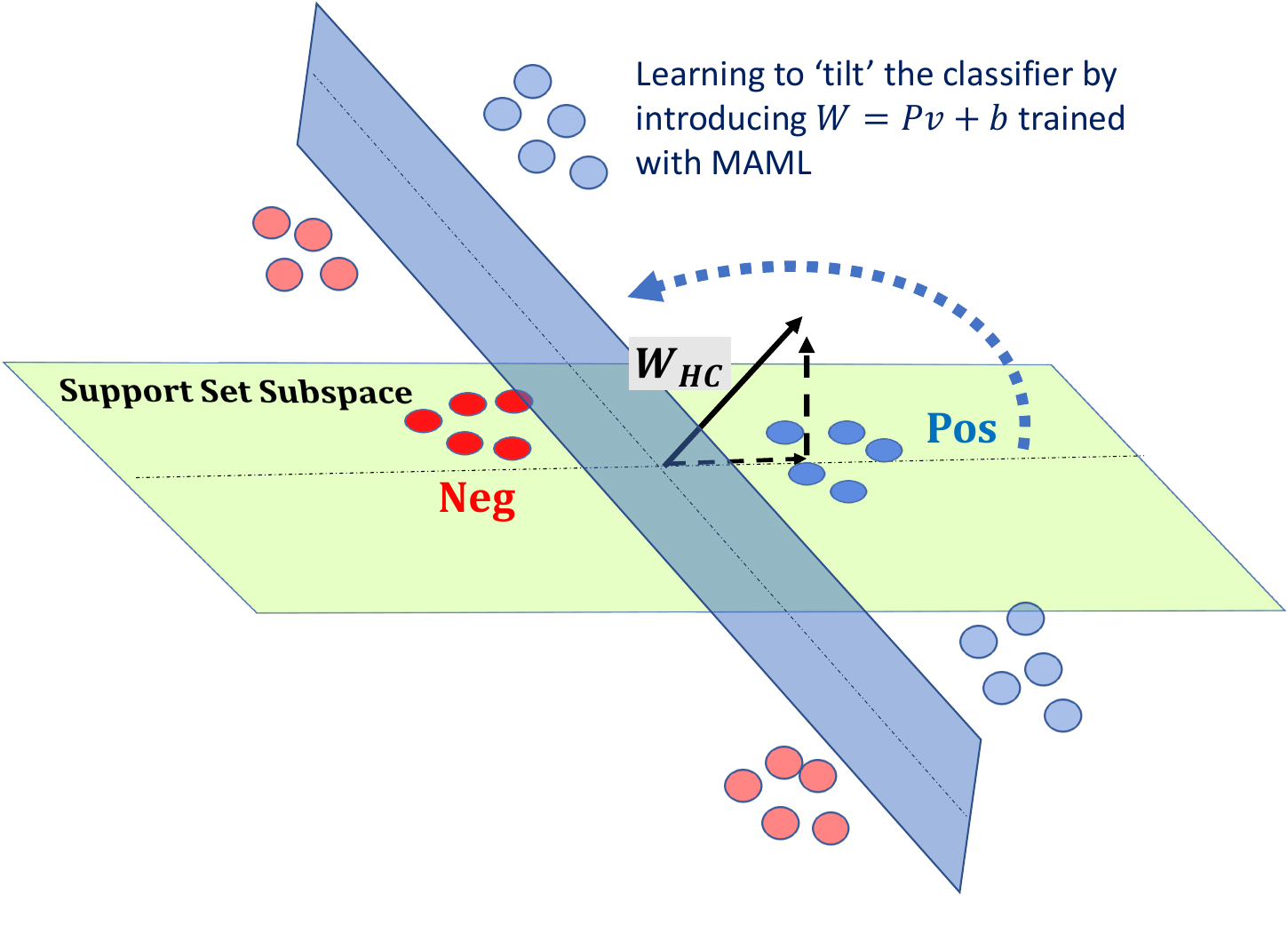}
    \caption{Hyper-Class Classifier}
    \label{fig:HC_Illustration}
    \end{subfigure}
    \caption{Illustration of HyperClass vs. common Linear Classifier for binary classification. (a) The few samples in the support set determine a subspace (green hyper-plane). Positive and negative samples are in blue and red respectively. Bold points represent samples in the support set and transparent points represent unseen samples from the query set. Standard Prototypical and linear classifier \eg Logistic Regression define classifiers that can rotate in the support plane (classifier vector $W$ is normal to the separating hyper-plane) and therefore remain with False-Positive (FP) and False Negative (FN) errors for samples out of the support set plane. (b) Our HyperClass method allows out-of plane rotation,  generalizing a classifier to unseen and out of (support) plane samples.}
    \label{fig:Subspace_Illustration}
\end{figure*}

Using the gradient terms in \eqref{eq:GradientsPvb}  we derive the update scheme $\forall x \in x^s$ for the first gradient step (again assuming unit learning rate, w.l.o.g):
\begin{align}
W_{1} &= P_{1}v_{1} + b_{1}  \\ \nonumber
& = (P_0 - \nabla_{P}\mathcal{L})(v_0 -\nabla_{v}\mathcal{L})+b_0-\nabla_{b}\mathcal{L} \\ \nonumber
&= (P_0+\lambda_x xv_0^T)(v_0+\lambda_x P_0^Tx) + b_0 +\lambda_x x \\ \nonumber
& = W_0 + \lambda_x P_0P_0^Tx +\lambda_x ||v_0||_{2}^{2}x \\ \nonumber  
& ~~~~~~~~~~~+ \lambda_x^2(v_0^TP_0^Tx)x +\lambda_x x.
\end{align}
Hereby we obtain the following update scheme as described in Eq. (8) in the paper:
\begin{equation}
    \Delta W = W_1 - W_0 = \lambda_x P_0P_0^Tx + Cx,
\end{equation}
with $C \in \mathbb{R}$ is a scalar value. This shows the impact of the projection head. While the second component in $\Delta W$ lies in the same support set subspace as in LR, the first term $\lambda_x PP^Tx^+$ can ''tilt" the classifier in a direction out of this {\it subspace}, extending the hypothesis class to a richer set of solutions. Note that the HyperClass is able to fuse knowledge from various tasks, locally (intra-task) and globally (inter-task) for better generalization. An illustration of this effect is shown in Fig. \ref{fig:Subspace_Illustration}, showing that this additional transformation (learned globally, over tasks) can predict an "out-of-plane" rotation of the decision boundary to reduce overfit when training only on few samples.

As we continue to compute the updates for $v, P, b$ in the next gradient steps we obtain:
\begin{align}
    \label{eq: v,p,b updates}
    & v_2 = \mathbb{S}(\{v_0, P_0^Tx\}) \subset \mathbb{R}^d \\ \nonumber
    & P_2 = \mathbb{S}(\{P_0,x v_0^T, \boldsymbol{x x^T P_0}\}) \subset \mathbb{R}^{d \times d} \\
    & b_2 = \mathbb{S}(\{b_0, x\}) \subset \mathbb{R}^d \nonumber.
\end{align}
where $\mathbb{S}(\cdot)$ denotes the span of a vector set. We are interested in the direction of the classifier $W$ in the embedding space, hence we omit the scalar values and consider only vector terms.
Note the fact that $v$ and $P$ are updated in outer-loop, thus they carry a {\it global} information (over various tasks), in contrast to $x$ that is $local$, within certain task. We further note that, in \eqref{eq: v,p,b updates}, a new term for the update of $P$ emerges: $xx^TP_0$ (indicated in bold).
From the Bayesian perspective, the task-adapted $P$ begins with prior $P_0$, learnt globally from previous tasks. The prior is then adapted with the second-order statistics of the evidence (task at hand), resulting in $xx^TP_0$. 

Further deriving the classifier after $k \geq 2$ gradient steps over the entire support set yields the following term, 
 \begin{align}
W_{k} = &W_0 + P_0P_0^T X^T \beta^1 + X^T \beta^2 + \beta^3 XX^TW_0 \nonumber \\  
& ~~~~~~+ XX^T(P_0P_0^T X^T) \beta^4,
\end{align}
denoted in matrix notation for sake of brevity. The vectors $\beta^1, \beta^2, \beta^4 \in \mathbb{R}^{|x^s|}$ are associated with the support samples (scalar for each sample) and $\beta^3 \in \mathbb{R}$ is a scalar. We observe that the new terms added are in fact the same terms previously introduced, weighted by the sample covariance matrix $XX^T$ (assuming zero-centered features), adding a second order statistics and a new source of information in determining the direction of $W$ in the embedding space.

One might also find the resemblance to PCA. PCA is an optimal linear dimensionality reduction (in $L_2$ sense) that achieves optimality whenever the exact covariance matrix is known. However, the optimal solution is often not achieved in cases where the covariance matrix is computed only from {\it few samples}, due to inaccuracy of the covariance matrix. One can find relevance to the well known problem of high-dimension, low-sample-size (HDLSS) data situations in statistics \cite{HDLSS2009,HDLSS2010}. In this sense, our method can be viewed as a method that modifies an unreliable covariance/correlation matrix by learning priors globally (in meta-learning).
\section{Sensitivity of Image Retrieval to Different Feedback Patterns}
\begin{figure*}[!ht]
    \centering
    \includegraphics[width=0.35\textwidth]{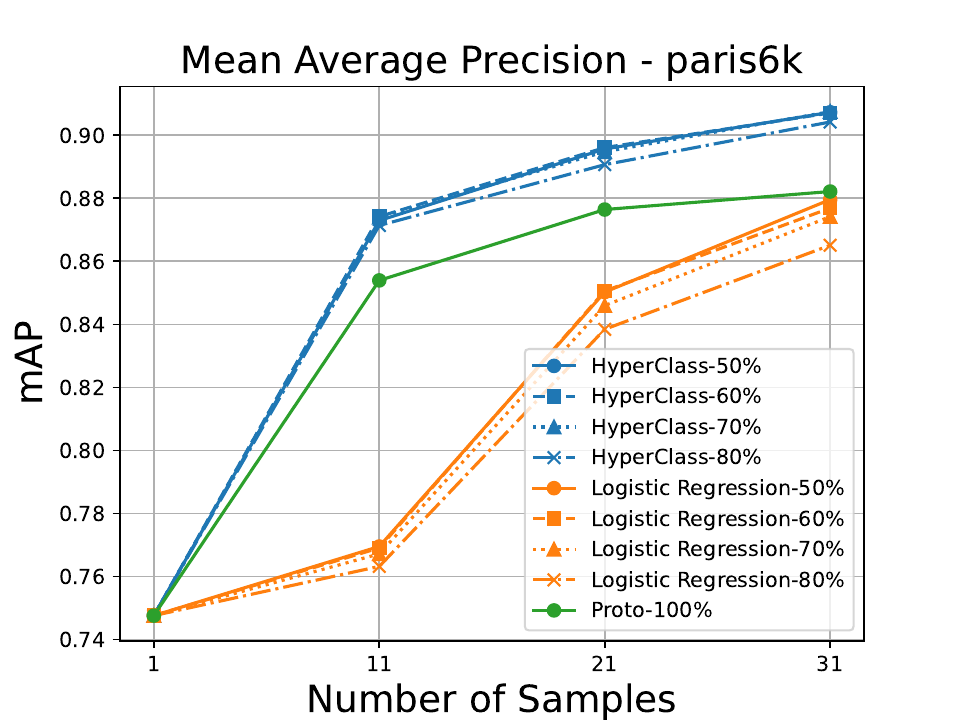}
    \includegraphics[width=0.35\textwidth]{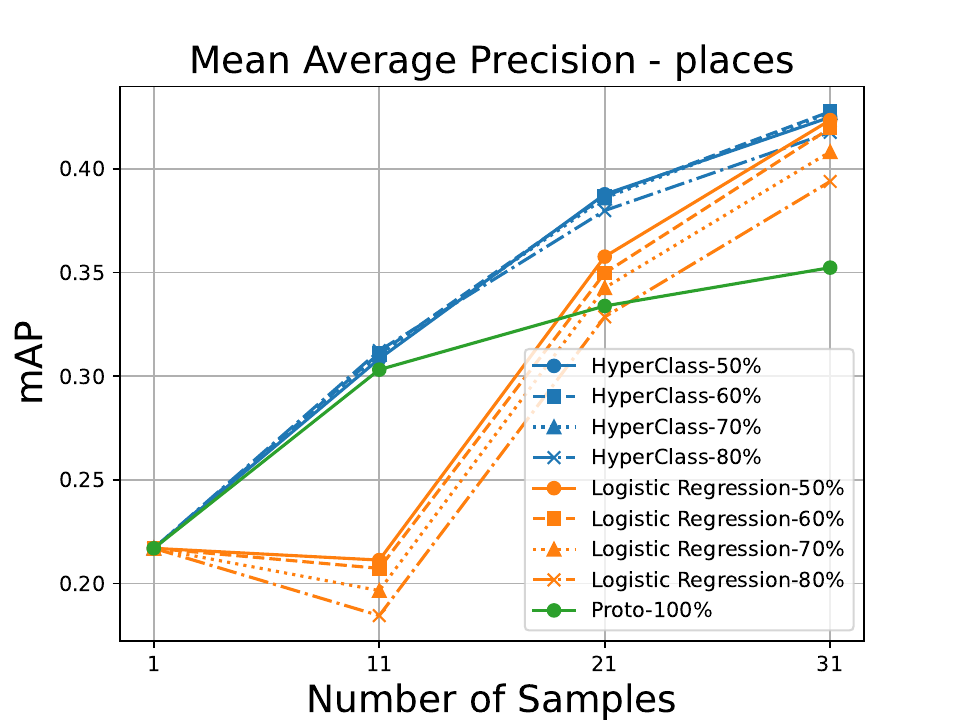}
    \includegraphics[width=0.35\textwidth]{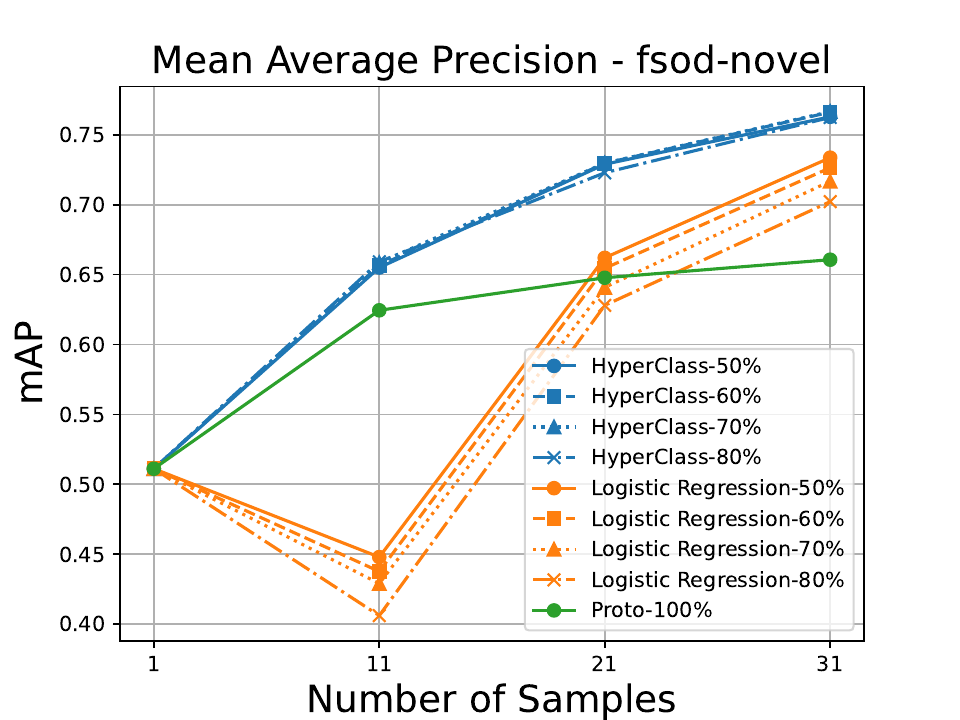}
    \includegraphics[width=0.35\textwidth]{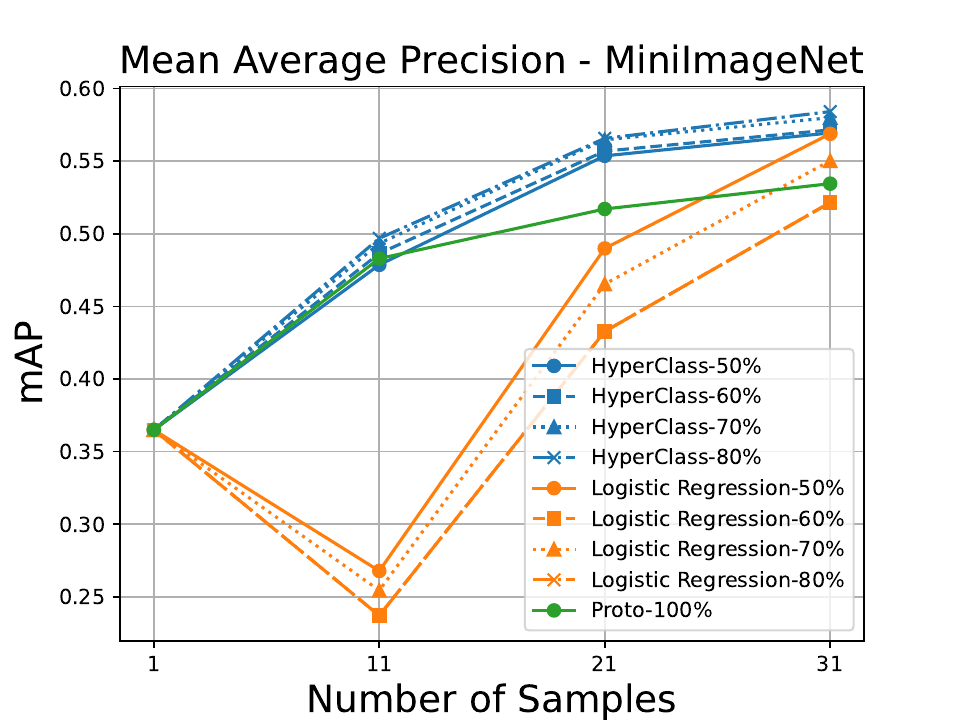}
    \caption{Mean Average Precision (mAP) on Image Retrieval with Relevance Feedback for various feedback selection patterns, from 50\%-50\% positive vs. negative to 80\% positive vs. 20\% negative. We compare our HyperClass to two alternatives, LR and Proto with 100\% positive samples (with the same number of train samples). X-axis denotes the number of samples available for training (few-shots). Note the insensitivity of HyperClass to different selection patterns.}
    \label{fig:FSIR-Different_Feedback_Ratios}
\end{figure*}

In the main paper we showed the results of IRRF task where 80\% of the selected samples in the feedback are positive and 20\% negative. Figure \ref{fig:FSIR-Different_Feedback_Ratios} shows the results for various selection patterns, changing from 50\%-50\% up to 80\%-20\%. The first key encouraging insight is that HyperClass is nearly invariant to this wide range of selection patterns. We further observe that HyperClass outperforms the comparing methods of Logistic Regression (LR) and even Proto approach (based on Prototypical approach) for all iterations and in all the benchmarks, regardless of the user selection pattern. This has a particular applicable impact, as positive samples are often rare in a large corpus and may be unavailable for feedback selection at initial rounds. Note the HyperClass superior performance over Proto, although we test Proto with more positive samples. 

\section{Rank Evolution Examples}
In this experiment we show the rank evolution of a certain relevant sample during the feedback cycles. The figure illustrates how relevant samples that were initially (searched with one query image) ranked over 100 in the list, propagate toward the top within 4 feedback cycles (obtaining 10 feedback tags per-cycle). Note that the names of the query object indicate the user's intent and are provided for sake of illustration here and they are not available to the model. Examples exhibit challenge as from the first glance, without knowing the object, the user's intent is unclear. The user's intent for "Goggles" could be perceived as a man's face, hat or even a pink jacket. For the "Sports uniform" it could be also a crowd. Yet, after accumulating several feedback samples the model is successful in finding the correct images and reveal the user's intent based on the common pattern, in the feedback images. 
\begin{figure*}[ht]
   \includegraphics[width=1\linewidth]{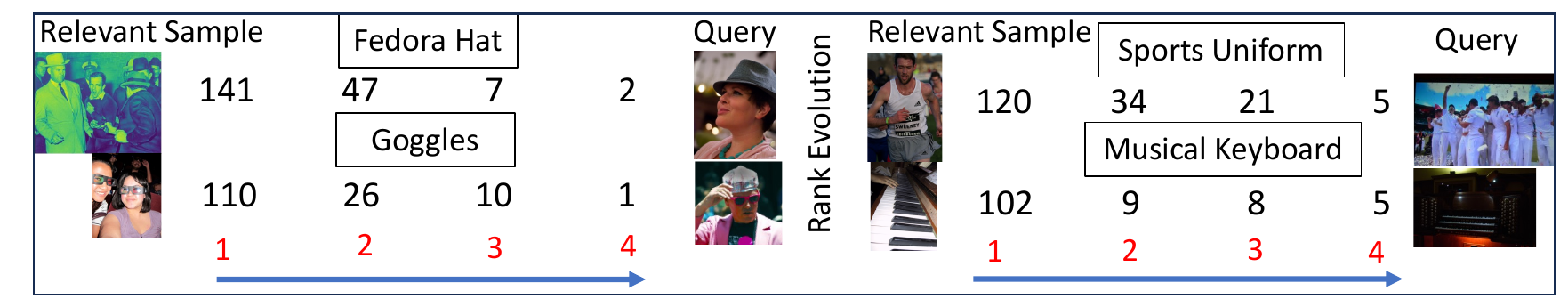}
   \caption{Rank evolution in several retrieved images during the feedback rounds (in red). The ranks at each round for the relevant images (at left) are presented, with the starting query images at right. The figure shows the gradual promotion of a relevant image in rank along the feedback rounds. The user's intent in text is provided for sake of illustration and is not available to the model. Note the ambiguity involved with the query image, as it can refer to search of various concepts in that image.
}
\end{figure*}

\section{More Implementation Details}
In this section we provide more details on our implementation for HyperClass. We follow ~\cite{GoodEmbedding} and use the same backbone of ResNet-12 ~\cite{resnet} for feature extraction in FSOCC and FSOR experiments, resulting in features dimension of $d=640$. For the Image Retrieval with Relevance Feedback experiments, we use ResNet-50 backbone ~\cite{resnet} pre-trained on Imagenet-21K \cite{imagenet21k}.
The outer and inner loop learning rates (LR) were 0.001 and 0.5 respectively, set by a grid search over (0.01, 0.001, 0.0001) and (0.1, 0.5, 1) on the validation set. We use 'learn2learn' package ~\cite{learn2learn} for the implementation of MAML with Adam ~\cite{adam} optimizer in the outer-loop iterations, applied with weight decay of 0.001. We train for 300 meta-batches, with 100 tasks per-batch, however in most cases the model converges much faster. Eventually, we select the best performing model (AP for image retrieval and AUROC for the rest) on the validation set. We update with 5 gradient steps in the inner loop (by vanilla SGD) and apply $l_2$ regularization on the parameters with weight of 1e-4 after a small search performed on the range (1e-2, 1e-3, 1e-4). We also use random crop, color jittering and horizontal flip augmentations as performed in ~\cite{GoodEmbedding} with 5 different augmentations for every image.

All of our experiments were run with single A100 GPU-40GB memory on Ubuntu 20.4 machine, each run taking around 3 hours.

For the implementation of Logistic Regression and One-Class SVM, we used the 'sklearn' package ~\cite{sklearn}.


\newpage

%% file: arxiv_version.bbl
\begin{thebibliography}{53}
\providecommand{\natexlab}[1]{#1}

\bibitem[{Alarc{\~a}o et~al.(2022)Alarc{\~a}o, Mendon{\c{c}}a, Maruta, and Fonseca}]{Expertoslf2022}
Alarc{\~a}o, S.~M.; Mendon{\c{c}}a, V.; Maruta, C.; and Fonseca, M.~J. 2022.
\newblock {ExpertosLF}: dynamic late fusion of CBIR systems using online learning with relevance feedback.
\newblock \emph{Multimedia Tools and Applications}, 1--43.

\bibitem[{Antoniou, Edwards, and Storkey(2018)}]{MAML++}
Antoniou, A.; Edwards, H.; and Storkey, A. 2018.
\newblock How to train your {MAML}.
\newblock \emph{arXiv:1810.09502}.

\bibitem[{Arnold et~al.(2020)Arnold, Mahajan, Datta, Bunner, and Zarkias}]{learn2learn}
Arnold, S.~M.; Mahajan, P.; Datta, D.; Bunner, I.; and Zarkias, K.~S. 2020.
\newblock {Learn2Learn}: A library for meta-learning research.
\newblock \emph{arXiv:2008.12284}.

\bibitem[{Banerjee et~al.(2018)Banerjee, Kurtz, Devorah, Do, Rubin, and Beaulieu}]{medical_relevanceFeedback2018}
Banerjee, I.; Kurtz, C.; Devorah, A.~E.; Do, B.; Rubin, D.~L.; and Beaulieu, C.~F. 2018.
\newblock Relevance feedback for enhancing content based image retrieval and automatic prediction of semantic image features: Application to bone tumor radiographs.
\newblock \emph{Journal of biomedical informatics}, 84: 123--135.

\bibitem[{Barz, K{\"a}ding, and Denzler(2018)}]{ITAL2018}
Barz, B.; K{\"a}ding, C.; and Denzler, J. 2018.
\newblock Information-theoretic active learning for content-based image retrieval.
\newblock In \emph{German Conference on Pattern Recognition}, 650--666. Springer.

\bibitem[{Chen et~al.(2019)Chen, Liu, Kira, Wang, and Huang}]{chen2019closer}
Chen, W.-Y.; Liu, Y.-C.; Kira, Z.; Wang, Y.-C.~F.; and Huang, J.-B. 2019.
\newblock A closer look at few-shot classification.
\newblock \emph{ICLR}.

\bibitem[{Chen et~al.(2021)Chen, Liu, Xu, Darrell, and Wang}]{MetaBaseline2021}
Chen, Y.; Liu, Z.; Xu, H.; Darrell, T.; and Wang, X. 2021.
\newblock Meta-baseline: Exploring simple meta-learning for few-shot learning.
\newblock In \emph{CVPR}, 9062--9071.

\bibitem[{Chowdhury et~al.(2021)Chowdhury, Jiang, Chaudhuri, and Jermaine}]{simpleFSL_ICCV21}
Chowdhury, A.; Jiang, M.; Chaudhuri, S.; and Jermaine, C. 2021.
\newblock Few-shot image classification: Just use a library of pre-trained feature extractors and a simple classifier.
\newblock In \emph{CVPR}, 9445--9454.

\bibitem[{Dang-Nguyen et~al.(2017)Dang-Nguyen, Piras, Giacinto, Boato, and Natale}]{dang2017multimodal}
Dang-Nguyen, D.-T.; Piras, L.; Giacinto, G.; Boato, G.; and Natale, F. G.~D. 2017.
\newblock Multimodal retrieval with diversification and relevance feedback for tourist attraction images.
\newblock \emph{ACM Transactions on Multimedia Computing, Communications, and Applications}, 13(4): 1--24.

\bibitem[{de~Ves et~al.(2016)de~Ves, Benavent, Coma, and Ayala}]{novelRF_neuroComputing2016}
de~Ves, E.; Benavent, X.; Coma, I.; and Ayala, G. 2016.
\newblock A novel dynamic multi-model relevance feedback procedure for content-based image retrieval.
\newblock \emph{Neurocomputing}, 208: 99--107.

\bibitem[{Deng et~al.(2009)Deng, Dong, Socher, Li, Li, and Fei-Fei}]{imagenet}
Deng, J.; Dong, W.; Socher, R.; Li, L.; Li, K.; and Fei-Fei, L. 2009.
\newblock Imagenet: A large-scale hierarchical image database.
\newblock In \emph{CVPR}, 248--255.

\bibitem[{Dhillon et~al.(2019)Dhillon, Chaudhari, Ravichandran, and Soatto}]{fs_baseline}
Dhillon, G.~S.; Chaudhari, P.; Ravichandran, A.; and Soatto, S. 2019.
\newblock A baseline for few-shot image classification.
\newblock \emph{arXiv:1909.02729}.

\bibitem[{Fan et~al.(2020)Fan, Zhuo, Tang, and Tai}]{fsod}
Fan, Q.; Zhuo, W.; Tang, C.-K.; and Tai, Y.-W. 2020.
\newblock Few-Shot Object Detection with Attention-{RPN} and Multi-Relation Detector.
\newblock In \emph{CVPR}.

\bibitem[{Finn, Abbeel, and Levine(2017)}]{MAML}
Finn, C.; Abbeel, P.; and Levine, S. 2017.
\newblock Model-agnostic meta-learning for fast adaptation of deep networks.
\newblock In \emph{International conference on machine learning}, 1126--1135. PMLR.

\bibitem[{Frikha et~al.(2021)Frikha, Krompa{\ss}, K{\"o}pken, and Tresp}]{fs_occ-AAAI}
Frikha, A.; Krompa{\ss}, D.; K{\"o}pken, H.-G.; and Tresp, V. 2021.
\newblock Few-shot one-class classification via meta-learning.
\newblock In \emph{AAAI}, volume~35, 7448--7456.

\bibitem[{Grandvalet and Bengio(2004)}]{entropy_minimization}
Grandvalet, Y.; and Bengio, Y. 2004.
\newblock Semi-supervised learning by entropy minimization.
\newblock \emph{Advances in neural information processing systems}, 17.

\bibitem[{He et~al.(2016)He, Zhang, Ren, and Sun}]{resnet}
He, K.; Zhang, X.; Ren, S.; and Sun, J. 2016.
\newblock Deep residual learning for image recognition.
\newblock In \emph{CVPR}, 770--778.

\bibitem[{Hu et~al.(2022)Hu, Li, St{\"u}hmer, Kim, and Hospedales}]{PMF_CVPR22}
Hu, S.~X.; Li, D.; St{\"u}hmer, J.; Kim, M.; and Hospedales, T.~M. 2022.
\newblock Pushing the limits of simple pipelines for few-shot learning: External data and fine-tuning make a difference.
\newblock In \emph{CVPR}, 9068--9077.

\bibitem[{Huang et~al.(2022)Huang, Ma, Han, and Chang}]{FSOR_cvpr2022}
Huang, S.; Ma, J.; Han, G.; and Chang, S.-F. 2022.
\newblock Task-Adaptive Negative Envision for Few-Shot Open-Set Recognition.
\newblock In \emph{CVPR}, 7171--7180.

\bibitem[{Jeong, Choi, and Kim(2021)}]{FSOR_cvpr2021}
Jeong, M.; Choi, S.; and Kim, C. 2021.
\newblock Few-shot open-set recognition by transformation consistency.
\newblock In \emph{CVPR}, 12566--12575.

\bibitem[{Jung and Marron(2009)}]{HDLSS2009}
Jung, S.; and Marron, J.~S. 2009.
\newblock PCA consistency in high dimension, low sample size context.
\newblock \emph{The Annals of Statistics}, 37(6B).

\bibitem[{Karamti et~al.(2018)Karamti, Tmar, Visani, Urruty, and Gargouri}]{OptimizedQueryIRRF2018}
Karamti, H.; Tmar, M.; Visani, M.; Urruty, T.; and Gargouri, F. 2018.
\newblock Vector space model adaptation and pseudo relevance feedback for content-based image retrieval.
\newblock \emph{Multimedia Tools and Applications}, 77: 5475--5501.

\bibitem[{Kingma and Ba(2014)}]{adam}
Kingma, D.~P.; and Ba, J. 2014.
\newblock Adam: A method for stochastic optimization.
\newblock \emph{arXiv:1412.6980}.

\bibitem[{Kozerawski and Turk(2021)}]{fs_occ-kozerawski}
Kozerawski, J.; and Turk, M. 2021.
\newblock One-Class Meta-Learning: Towards Generalizable Few-Shot Open-Set Classification.
\newblock \emph{arXiv:2109.06859}.

\bibitem[{Kruspe(2019)}]{1way_proto}
Kruspe, A. 2019.
\newblock One-way prototypical networks.
\newblock \emph{arXiv:1906.00820}.

\bibitem[{Lee et~al.(2019{\natexlab{a}})Lee, Maji, Ravichandran, and Soatto}]{MetaOptNet_CVPR19}
Lee, K.; Maji, S.; Ravichandran, A.; and Soatto, S. 2019{\natexlab{a}}.
\newblock Meta-learning with differentiable convex optimization.
\newblock In \emph{CVPR}, 10657--10665.

\bibitem[{Lee et~al.(2019{\natexlab{b}})Lee, Maji, Ravichandran, and Soatto}]{differential_svm}
Lee, K.; Maji, S.; Ravichandran, A.; and Soatto, S. 2019{\natexlab{b}}.
\newblock Meta-learning with differentiable convex optimization.
\newblock In \emph{CVPR}, 10657--10665.

\bibitem[{Lin(2022)}]{lin2022block}
Lin, W.-C. 2022.
\newblock Block-based pseudo-relevance feedback for image retrieval.
\newblock \emph{Journal of Experimental \& Theoretical Artificial Intelligence}, 34(5): 891--903.

\bibitem[{Liu et~al.(2020)Liu, Kang, Li, Hua, and Vasconcelos}]{FSOR_cvpr2020}
Liu, B.; Kang, H.; Li, H.; Hua, G.; and Vasconcelos, N. 2020.
\newblock Few-shot open-set recognition using meta-learning.
\newblock In \emph{CVPR}, 8798--8807.

\bibitem[{Liu et~al.(2017)Liu, Feng, Liu, Wu, Sun, and Wang}]{liu2017robust}
Liu, S.; Feng, L.; Liu, Y.; Wu, J.; Sun, M.; and Wang, W. 2017.
\newblock Robust discriminative extreme learning machine for relevance feedback in image retrieval.
\newblock \emph{Multidimensional Systems and Signal Processing}, 28: 1071--1089.

\bibitem[{MacArthur et~al.(2002)MacArthur, Brodley, Kak, and Broderick}]{IIR_CVIU02}
MacArthur, S.~D.; Brodley, C.~E.; Kak, A.~C.; and Broderick, L.~S. 2002.
\newblock Interactive content-based image retrieval using relevance feedback.
\newblock \emph{Computer Vision and Image Understanding}, 88(2): 55--75.

\bibitem[{Mehra, Hamm, and Belkin(2018)}]{interactive_CBIR_2018}
Mehra, A.; Hamm, J.; and Belkin, M. 2018.
\newblock Fast Interactive Image Retrieval using large-scale unlabeled data.
\newblock \emph{arXiv:1802.04204}.

\bibitem[{Ngo, Ngo, and Nguyen(2016)}]{interactive_CBIR_2016}
Ngo, G.~T.; Ngo, T.~Q.; and Nguyen, D.~D. 2016.
\newblock Image retrieval with relevance feedback using SVM active learning.
\newblock \emph{International Journal of Electrical and Computer Engineering}, 6(6): 3238.

\bibitem[{Pedregosa et~al.(2011)Pedregosa, Varoquaux, Gramfort, Michel, Thirion, Grisel, Blondel, Prettenhofer, Weiss, Dubourg et~al.}]{sklearn}
Pedregosa, F.; Varoquaux, G.; Gramfort, A.; Michel, V.; Thirion, B.; Grisel, O.; Blondel, M.; Prettenhofer, P.; Weiss, R.; Dubourg, V.; et~al. 2011.
\newblock Scikit-learn: Machine learning in Python.
\newblock \emph{the Journal of machine Learning research}, 12: 2825--2830.

\bibitem[{Pinjarkar, Sharma, and Selot(2018)}]{pinjarkar2020deep}
Pinjarkar, L.; Sharma, M.; and Selot, S. 2018.
\newblock Deep {CNN} combined with relevance feedback for trademark image retrieval.
\newblock \emph{Journal of Intelligent Systems}, 29(1): 894--909.

\bibitem[{Putzu, Piras, and Giacinto(2020)}]{putzu2020convolutional}
Putzu, L.; Piras, L.; and Giacinto, G. 2020.
\newblock Convolutional neural networks for relevance feedback in content based image retrieval.
\newblock \emph{Multimedia Tools and Applications}, 79: 26995--27021.

\bibitem[{Qazanfari, Hassanpour, and Qazanfari(2017)}]{RelevanceFeedback_ICSPIS2017}
Qazanfari, H.; Hassanpour, H.; and Qazanfari, K. 2017.
\newblock A short-term learning framework based on relevance feedback for content-based image retrieval.
\newblock In \emph{2017 3rd Iranian Conference on Intelligent Systems and Signal Processing (ICSPIS)}, 136--140. IEEE.

\bibitem[{Radenovi{\'c} et~al.(2018)Radenovi{\'c}, Iscen, Tolias, Avrithis, and Chum}]{new_paris_oxford}
Radenovi{\'c}, F.; Iscen, A.; Tolias, G.; Avrithis, Y.; and Chum, O. 2018.
\newblock Revisiting oxford and paris: Large-scale image retrieval benchmarking.
\newblock In \emph{CVPR}, 5706--5715.

\bibitem[{Rao et~al.(2018)Rao, Liu, Fan, Song, and Yang}]{SVM_IRRF2018}
Rao, Y.; Liu, W.; Fan, B.; Song, J.; and Yang, Y. 2018.
\newblock A novel relevance feedback method for {CBIR}.
\newblock \emph{World Wide Web}.

\bibitem[{Ravi and Larochelle(2017)}]{ravi2017optimization}
Ravi, S.; and Larochelle, H. 2017.
\newblock Optimization as a model for few-shot learning.
\newblock In \emph{ICML}.

\bibitem[{Ren et~al.(2018)Ren, Triantafillou, Ravi, Snell, Swersky, Tenenbaum, Larochelle, and Zemel}]{tieredImageNet}
Ren, M.; Triantafillou, E.; Ravi, S.; Snell, J.; Swersky, K.; Tenenbaum, J.~B.; Larochelle, H.; and Zemel, R.~S. 2018.
\newblock Meta-learning for semi-supervised few-shot classification.
\newblock \emph{ICML}.

\bibitem[{Ridnik et~al.(2021)Ridnik, Ben-Baruch, Noy, and Zelnik-Manor}]{imagenet21k}
Ridnik, T.; Ben-Baruch, E.; Noy, A.; and Zelnik-Manor, L. 2021.
\newblock Imagenet-21k pretraining for the masses.
\newblock \emph{arXiv:2104.10972}.

\bibitem[{Rodr{\'\i}guez et~al.(2020)Rodr{\'\i}guez, Laradji, Drouin, and Lacoste}]{EPFNET_ECCV20}
Rodr{\'\i}guez, P.; Laradji, I.; Drouin, A.; and Lacoste, A. 2020.
\newblock Embedding propagation: Smoother manifold for few-shot classification.
\newblock In \emph{ECCV}, 121--138.

\bibitem[{Satish and Supreethi(2017)}]{BayesianNetworkClassifierIRRF}
Satish, B.; and Supreethi, K. 2017.
\newblock Content based medical image retrieval using relevance feedback Bayesian network.
\newblock In \emph{2017 International Conference on Electrical, Electronics, Communication, Computer, and Optimization Techniques}, 424--430. IEEE.

\bibitem[{Snell, Swersky, and Zemel(2017)}]{proto_net}
Snell, J.; Swersky, K.; and Zemel, R. 2017.
\newblock Prototypical networks for few-shot learning.
\newblock \emph{Advances in neural information processing systems}, 30.

\bibitem[{Tian et~al.(2020)Tian, Wang, Krishnan, Tenenbaum, and Isola}]{GoodEmbedding}
Tian, Y.; Wang, Y.; Krishnan, D.; Tenenbaum, J.~B.; and Isola, P. 2020.
\newblock Rethinking few-shot image classification: a good embedding is all you need?
\newblock In \emph{ECCV}, 266--282.

\bibitem[{Vinyals et~al.(2016)Vinyals, Blundell, Lillicrap, Wierstra et~al.}]{miniImagenet}
Vinyals, O.; Blundell, C.; Lillicrap, T.; Wierstra, D.; et~al. 2016.
\newblock Matching networks for one shot learning.
\newblock \emph{Advances in neural information processing systems}, 29.

\bibitem[{Wang et~al.(2016)Wang, Liang, Li, Li, and Yang}]{wang2016new}
Wang, X.-Y.; Liang, L.-L.; Li, W.-Y.; Li, D.-M.; and Yang, H.-Y. 2016.
\newblock A new {SVM}-based relevance feedback image retrieval using probabilistic feature and weighted kernel function.
\newblock \emph{Journal of Visual Communication and Image Representation}, 38: 256--275.

\bibitem[{Xu, Wang, and Mao(2017)}]{OptimizedQueryIRRF2017}
Xu, H.; Wang, J.-y.; and Mao, L. 2017.
\newblock Relevance feedback for Content-based Image Retrieval using deep learning.
\newblock In \emph{ICIVC}, 629--633. IEEE.

\bibitem[{Yata and Aoshima(2010)}]{HDLSS2010}
Yata, K.; and Aoshima, M. 2010.
\newblock Effective PCA for high-dimension, low-sample-size data with singular value decomposition of cross data matrix.
\newblock \emph{Journal of multivariate analysis}, 101(9): 2060--2077.

\bibitem[{Ye et~al.(2020)Ye, Hu, Zhan, and Sha}]{fs_transformer}
Ye, H.-J.; Hu, H.; Zhan, D.-C.; and Sha, F. 2020.
\newblock Few-shot learning via embedding adaptation with set-to-set functions.
\newblock In \emph{CVPR}, 8808--8817.

\bibitem[{Zhang et~al.(2020)Zhang, Cai, Lin, and Shen}]{DeepEMD_CVPR20}
Zhang, C.; Cai, Y.; Lin, G.; and Shen, C. 2020.
\newblock {DeepEMD: Few-shot image classification with differentiable earth mover's distance and structured classifiers}.
\newblock In \emph{CVPR}, 12203--12213.

\bibitem[{Zhou et~al.(2017)Zhou, Lapedriza, Khosla, Oliva, and Torralba}]{places}
Zhou, B.; Lapedriza, A.; Khosla, A.; Oliva, A.; and Torralba, A. 2017.
\newblock Places: A 10 million image database for scene recognition.
\newblock \emph{IEEE transactions on pattern analysis and machine intelligence}, 40(6): 1452--1464.

\end{thebibliography}
